\documentclass[a4paper, 10pt, conference]{ieeeconf}      
\usepackage{FG2025}
\usepackage{times}
\usepackage{epsfig}
\usepackage{graphicx}
\usepackage{amsmath}
\usepackage{amssymb}

\usepackage{booktabs}
\usepackage{commath}
\usepackage{mathtools}
\usepackage{enumerate}
\usepackage{xcolor}

\usepackage{caption}
\usepackage{subcaption}
\usepackage[breaklinks,colorlinks]{hyperref}

\usepackage{soul}

\usepackage[ruled,vlined]{algorithm2e}

\definecolor{amethyst}{rgb}{0.6, 0.4, 0.8}
\definecolor{mygray}{gray}{0.5}

\IEEEoverridecommandlockouts                              
\def\FGPaperID{0008} 

\title{\LARGE \bf
Exploiting Aggregation and Segregation of Representations for Domain Adaptive Human Pose Estimation
}

\author{\parbox{16cm}{\centering
    {\large Qucheng Peng$^1$, Ce Zheng$^2$, Zhengming Ding$^3$, Pu Wang$^4$, and Chen Chen$^1$}\\
    {\normalsize
    $^1$ Center for Research in Computer Vision, University of Central Florida, Orlando, USA\\
    $^2$Robotics Institute, Carnegie Mellon University, Pittsburgh, USA\\
    $^3$Department of Computer Science, Tulane University, New Orleans, USA\\
    $^4$University of North Carolina at Charlotte, Charlotte, USA}}
    \\
    {\tt\small \{qucheng.peng, chen.chen\}@ucf.edu}\\
}

\begin{document}

\FGfinalcopy
\thispagestyle{empty}
\pagestyle{empty}
\maketitle

\begin{abstract}

Human pose estimation (HPE) has received increasing attention recently due to its wide application in motion analysis, virtual reality, healthcare, etc. However, it suffers from the lack of labeled diverse real-world datasets due to the time- and labor-intensive annotation. To cope with the label deficiency issue, one common solution is to train the HPE models with easily available synthetic datasets (source) and apply them to real-world data (target) through domain adaptation (DA). Unfortunately, prevailing domain adaptation techniques within the HPE domain remain predominantly fixated on effecting alignment and aggregation between source and target features, often sidestepping the crucial task of excluding domain-specific representations. To rectify this, we introduce a novel framework that capitalizes on both representation aggregation and segregation for domain adaptive human pose estimation. Within this framework, we address the network architecture aspect by disentangling representations into distinct domain-invariant and domain-specific components, facilitating aggregation of domain-invariant features while simultaneously segregating domain-specific ones. Moreover, we tackle the discrepancy measurement facet by delving into various keypoint relationships and applying separate aggregation or segregation mechanisms to enhance alignment. Extensive experiments on various benchmarks, e.g., Human3.6M, LSP, H3D, and FreiHand, show that our method consistently achieves state-of-the-art performance. The project is available at \url{https://github.com/davidpengucf/EPIC}.

\end{abstract}

\section{Introduction}
\label{sec:intro}

2D human pose estimation (HPE) from monocular images is an important task in computer vision and has witnessed great success thanks to the advancement of deep neural networks.  
As with any task, the training of deep learning models necessitates vast amounts of labeled data  \cite{deng2009imagenet,ionescu2013human3,lin2014microsoft,zheng2023feater}. For the HPE task, collecting real-world datasets with diverse appearances and poses and generating ground-truth annotations are time- and labor-intensive. On the contrary, synthetic datasets with annotations are much cheaper and easier to acquire due to the development of computer graphics and game engines. Therefore, one can essentially generate an (infinite) number of labeled synthetic samples to train HPE models. However, models trained on synthetic data may not generalize well on real-world data in terms of the data distribution discrepancy. In this case, domain adaptation (DA) plays a key role in generalizing the synthetic data (source) trained models to real-world (target) data \cite{ben2010theory,zhuang2020comprehensive,peng2023rain, xin2024vpetl, peng2024dual, zheng20213d}. 

\begin{figure}[t]
  \centering
   \includegraphics[width=1.0\linewidth]{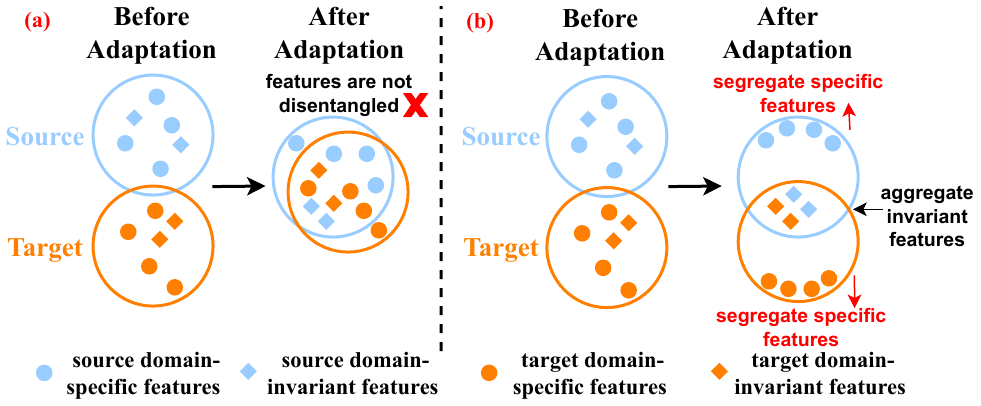}
   \caption{Comparison of (a) previous works and (b) our objective. Previous works align source and target directly, which may result in suboptimal performances due to the mixing of diverse features. Our objective aggregates domain-invariant features and segregates domain-specific features simultaneously to craft a good regressor in the target domain.
   }\vspace{-10pt}
   \label{fig:com_fra}
\end{figure}

In the context of DA, learned representations encompass two vital components: domain-invariant and domain-specific elements \cite{stojanov2021domain,gu2019improving}. This dichotomy is also observed in DA for HPE \cite{jiang2021regressive,jin2022branch,han2022transpar}. \textbf{Domain-invariant features}, such as kinematic attributes within human poses, enhance knowledge transfer across domains, while \textbf{domain-specific features} like body scale and viewpoint impede effective cross-domain learning. Conventional approaches \cite{mu2020learning,li2021synthetic,jiang2021regressive,jin2022branch,kim2022unified} in DA for HPE often amalgamate and align source and target features without discerning these intrinsic components (see Fig. \ref{fig:com_fra}a). These approaches often result in suboptimal performances due to the mixing of diverse features. To disentangle the mixed features better, \emph{we emphasize the incorporation of source-invariant and target-invariant features in the adaptation process, while preserving target-specific characteristics and reducing the impact of source-specific attributes.} As depicted in Fig. \ref{fig:com_fra}b, our approach aims to achieve this objective by disentangling representations from both the source and target domains into domain-invariant and domain-specific components. We prioritize target pose estimation by aggregating domain-invariant attributes, thus fostering positive knowledge transfer from source to target. Furthermore, we segregate source-specific and target-specific features, thereby shielding target-specific representations and mitigating any adverse effects stemming from source-specific representations.

\begin{figure}[t]
  \centering
   \includegraphics[width=1.0\linewidth]{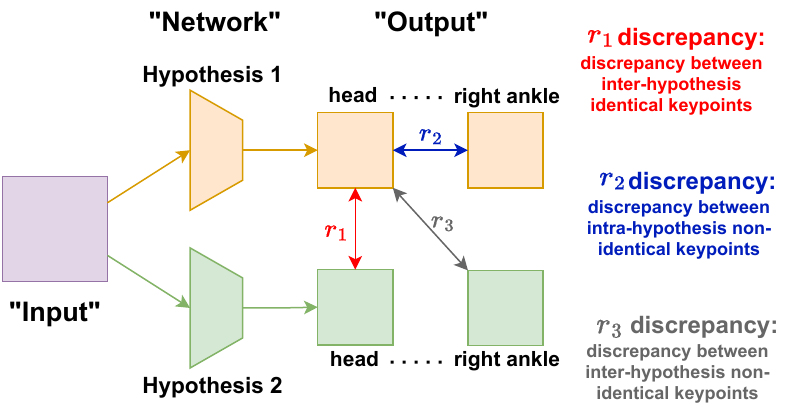}
   \caption{Hypothesis discrepancy based on multiple relations.
Apart from the $r_1$ discrepancy (marked \textcolor{red}{red}) that is considered by existing methods, we enhance identical hypotheses consistency by $r_2$
discrepancy (marked \textcolor{blue}{blue}) and non-identical hypotheses discrimination by $r_3$ discrepancy (marked \textcolor{mygray}{gray}).
   }%
   \vspace{-8pt}
   \label{fig:relation}
\end{figure}

While our previous discussion viewed aggregation and segregation from a network perspective, this is insufficient to tackle the complexities inherent in human pose estimation, whose output comprises multiple keypoints such as head, hips, and ankles (see Fig. \ref{fig:relation}), each with distinct attributes. For heatmaps corresponding to the same keypoint (keypoint-wise) or generated by the same hypothesis (hypothesis-wise), it is anticipated that they exhibit similarity in their representations through specific aggregation strategies, and conversely, dissimilarity should be appropriately measured through specific segregation strategies. To fulfill this need, we introduce a novel discrepancy measurement, depicted in Fig. \ref{fig:relation}. This measurement takes into account three distinct types of relations between pairs of estimation outputs: $r_1$ relations involving inter-hypothesis identical keypoints, $r_2$ relations encompassing intra-hypothesis non-identical keypoints, and $r_3$ relations associated with inter-hypothesis non-identical keypoints. Prevailing methodologies \cite{jiang2021regressive,kim2022unified,jin2022branch,han2022transpar} predominantly focus on aggregating $r_1$ relations exclusively. In contrast, our approach transcends these limitations. It not only aggregates $r_1$ relations but also incorporates aggregation of $r_2$ relations while simultaneously segregating $r_3$ relations. Detailed explanations are provided in Sec. \ref{sec:dl}. The primary contributions of this paper can be summarized as follows:

\begin{itemize}
    \item We introduce a novel framework that disentangles features into domain-invariant and domain-specific components, conducting aggregation and segregation on them separately to reduce the domain gap. 
    \item We propose a new hypothesis discrepancy measurement for human pose estimation, aggregating the intra-hypothesis non-identical discrepancy and segregating the inter-hypothesis non-identical discrepancy.
    \item We conduct comprehensive experiments on several domain-adaptive HPE benchmarks and the results show that our method outperforms the state-of-the-art approaches by a considerable margin. 
\end{itemize}

\section{Related Work}
\label{sec:related}

\subsection{Human Pose Estimation} 

In this paper, we focus on domain adaptation in 2D HPE, which can be divided into regression methods and heatmap-based methods. Regression methods directly output keypoints coordinates. Starting from \cite{toshev2014deeppose}, deep neural networks are applied in HPE. \cite{sun2017compositional} introduces a structure-aware regression method that uses bone-based representations. \cite{li2021pose} combines vision transformers \cite{dosovitskiy2020image}, while \cite{shi2022end} proposes a fully end-to-end method trained in one single stage. For heatmap-based methods, the networks first obtain heatmaps constructed by putting 2D Gaussian kernels on the potential keypoints, then convert heatmaps to coordinates. \cite{chen2018cascaded} uses global and refine nets to improve performance. \cite{xiao2018simple} proposes Simple Baselines to simplify the network structures. \cite{sun2019deep} fully utilizes all the features at different scales. Besides, some work tends to improve the quality of heatmaps, like \cite{zhang2020distribution} that extends Gaussian to second order in heatmaps and \cite{luo2021rethinking} that proposes scale-adaptive heatmaps. In our model, we follow previous works \cite{jiang2021regressive,kim2022unified} to use the heatmap-based approach and apply Simple Baselines as the backbone. 

\subsection{Domain Adaptation}

There are two kinds of approaches for domain adaptive image classification, i.e., statistics-matching and adversarial-based methods. Statistics-matching methods measure the discrepancy between source and target, and then minimize the discrepancy. To name a few, \cite{long2015learning, xin2024mmap} uses maximum mean discrepancy \cite{peng2022multi}, \cite{tzeng2014deep} applies deep domain confusion and \cite{sun2016deep, pinyoanuntapong2023gaitsada} utilizes correlation relation. \cite{wu2021heter} uses this paradigm for heterogeneous domain adaptation, and \cite{deng2021cluster} jointly makes clustering and discrimination for alignment. Adversarial-based methods are inspired by generative adversarial networks \cite{goodfellow2020generative}, which play a min-max game for two domains. For example, \cite{ganin2015unsupervised} builds confusion domains. \cite{saito2018maximum} tackles with the game of two classifiers. \cite{xu2019larger} shows that learning larger feature norms matter and \cite{tang2020unsupervised} reveals that discriminative clustering can benefit the adaptation. Our method applies the adversarial structure and also adopts the maximum mean discrepancy technique from statistics-matching.

\subsection{Domain Adaptive Human Pose Estimation}

Domain-adaptive HPE methods can be divided into two categories. One is the shared structure, in which the networks of source pretrain and target adaptation share weights. CC-SSL \cite{mu2020learning} uses one network trained in an end-to-end fashion. RegDA \cite{jiang2021regressive} applies one shared feature extractor and two separate regressors. TransPar \cite{han2022transpar} adopts a similar structure but focuses on transferable parameters. The other adopts an unshared structure, where the pretrain and adaptation form a teacher-student paradigm. To name a few, MDAM \cite{li2021synthetic} tackles this structure, together with a novel pseudo-label strategy. UniFrame \cite{kim2022unified} modifies the classic Mean-Teacher \cite{tarvainen2017mean} model, combining with style transfer \cite{huang2017arbitrary}. MarsDA \cite{jin2022branch} edits RegDA in a teacher-student manner \cite{peng20243d}. \cite{peng2023source, raychaudhuri2023prior} discuss a specific setting called ``Source-free'', which is different from others, so we do not compare to them.

The shared structure has fewer parameters and is easy to converge, but the shared weights suffer from domain shifts. The unshared one can resist domain shifts by separate networks, but the convergence speed is slower. Our method adopts the shared structure and uses intermediate representations to combat domain shifts.

\section{Methodology}

\subsection{Problem Statement}

In 2D HPE, we have a labeled pose dataset $\mathcal{D} = \{(x_{i},y_{i})\}$, where $x_{i} \in \mathbb{R}^{C \times H \times W}$ is the image and $y_{i} \in \mathbb{R}^{K\times 2}$ is the corresponding keypoint coordinates. Here $H$ and $W$ are the height and width of the image and $C$ is the number of channels. Besides, $K$ is the number of keypoints. For the sake of smoother training, most existing 2D methods use heatmaps $M_{i} \in \mathbb{R}^{K \times H' \times W'}$ to represent coordinates in a spatial way during supervised learning, where $H'$ and $W'$ are the height and width of heatmaps. In the inference stage, well-trained supervised models output heatmaps that predict the probability of joints occurring at each pixel. For each heatmap, the pixel with the highest probability is treated as the predicted joint's location. We define the transform from heatmaps to coordinates as $T: \mathbb{R}^{K \times H'\times W'} \rightarrow \mathbb{R}^{K\times 2}$, and the inverse transform from coordinates to heatmaps as $T^{-1}: \mathbb{R}^{K\times 2} \rightarrow \mathbb{R}^{K \times H'\times W'}$.

In the domain adaptive HPE setting, we have a source domain dataset $\mathcal{S} = \{(x^{s}_{i},y^{s}_{i})\}^{n_{s}}_{i=1}$ with $n_{s}$ labeled pose samples. $x^{s}_{i} \in \mathbb{R}^{C \times H \times W}$ is the source image and $y^{s}_{i} \in \mathbb{R}^{K\times 2}$ is the corresponding pose annotation. Besides, there exists a target domain dataset $\mathcal{T} = \{x^{t}_{i}\}^{n_{t}}_{i=1}$ that includes $n_{t}$ unlabeled pose samples. The source and target domains share the same label space but lie in different distributions. \textit{Our goal is to use the labeled source dataset and unlabeled target dataset to obtain a model that achieves high performance on the target dataset. }

\begin{figure}[t]
  \centering
   \includegraphics[width=1.0\linewidth]{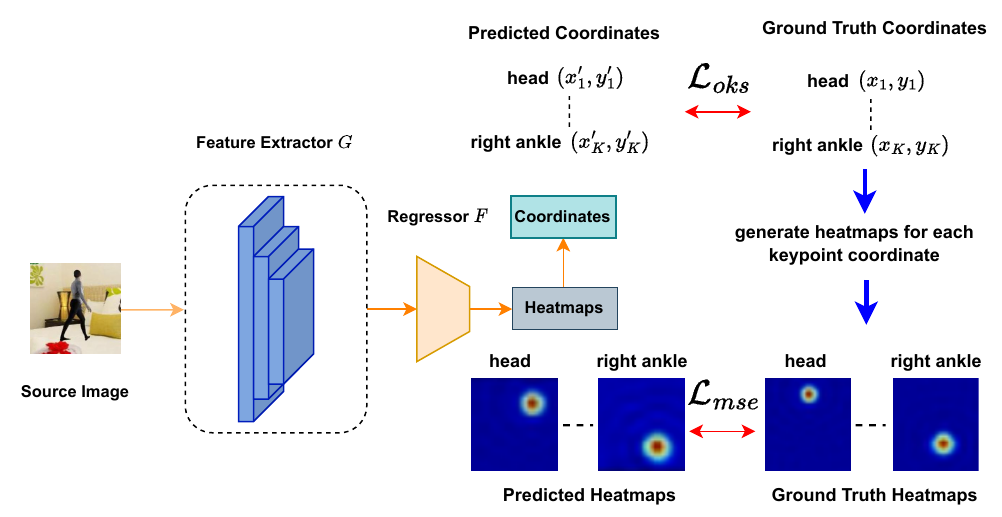}

   \caption{Pipeline of the supervised 2D HPE, and the \textbf{source-pretraining process} of our method. After passing feature extractor $G$ and regressor $F$, each source image turns to be $K$ heatmaps corresponding to $K$ keypoints. After certain transforms, $K$ 2D keypoints are obtained. We use these heatmaps and coordinates to compute $\mathcal{L}_{mse}$ and $\mathcal{L}_{oks}$. 
   }%
   \vspace{-10pt}
   \label{fig:pretrain}
\end{figure}

\subsection{Source Domain Pretraining}

For DA in HPE, labeled source data are used for model pretraining as the first step. Same as most existing domain adaptive pose estimation methods \cite{mu2020learning,li2021synthetic,jiang2021regressive,kim2022unified}, the network is composed of a feature extractor $G$ (such as ResNet backbone) and a pose regressor $F$ as shown in Fig. \ref{fig:pretrain}. Here we first adopt the coordinate loss called Objective Keypoint Similarity (OKS) loss \cite{shi2022end}, which can be represented as:  

\begin{equation}
  \mathcal{L}_{oks}(\hat{y},y) = \sum_{i=1}^{K} \exp\Big(\frac{-\norm{\hat{y}_{i} - y_{i}}_{\ell_2}}{2sk_{i}}\Big),
  \label{eq:oks}
\end{equation}
where $\hat{y}$ is the predicted coordinate and $y$ is the ground-truth coordinate. $\norm{\cdot}_{\ell_2}$ is the Euclidean distance between the prediction and ground truth of keypoint $i$, and $k_{i}$ is a keypoint constant that controls falloff. $s$ is the area of the input image. During the pretraining, we combine the heatmap-based loss (MSE loss) and the coordinate-based loss (OKS loss) as the overall loss:
\begin{equation}
\begin{multlined}
    \mathcal{L}_{pretrain} = \mathbb{E}_{(x^{s}_{i},y^{s}_{i}) \in \mathcal{S}} \mathcal{L}_{mse}(F(G(x_{i}^{s})),T^{-1}(y_{i}^{s})) \\
    + \mathbb{E}_{(x^{s}_{i},y^{s}_{i})\in \mathcal{S}} \mathcal{L}_{oks}(T(F(G(x_{i}^{s}))),y_{i}^{s}),
\end{multlined}
\label{eq:pretrain}
\end{equation}
where $\mathcal{L}_{mse}$ denotes the Mean Squared Error (MSE) loss between the heatmap representations of prediction and ground truth. Though the argmax operation in $T^{-1}$ is non-differentiable, we use a differentiable \emph{soft-argmax} proposed in \cite{sun2018integral}. 

\subsection{Network Structure for Domain Adaption}
\label{sec:net}

\begin{figure}[t]
  \centering
\hspace{-4mm}   \includegraphics[width=1.0\linewidth]{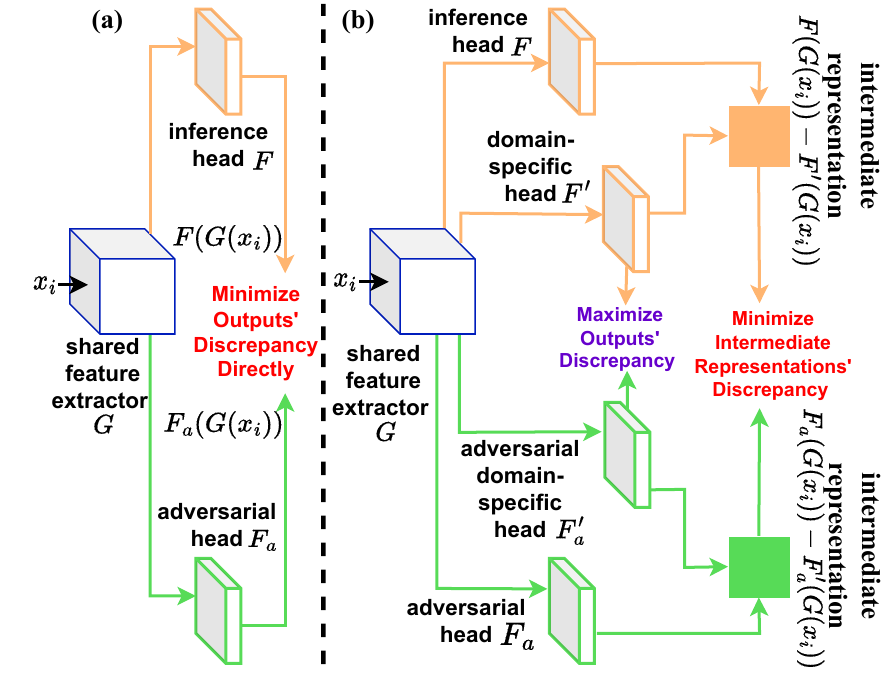}
   \caption{Comparisons between (a) conventional structure and (b) our proposed Intermediate Domain Framework (IDF). Details are provided in Section \ref{sec:net}. (Best viewed in color and zoom in)}%
   \vspace{-10pt}
   \label{fig:struc}
\end{figure}

In previous works \cite{jiang2021regressive,han2022transpar,jin2022branch}, the conventional network structure consists of one shared feature extractor $G$ and two regression heads: the inference head $F$ and the adversarial head $F_a$, as depicted in Fig. \ref{fig:struc}(a). The discrepancy measurement between these two heads is defined as $\norm{F(G(x_{i}))-F_a(G(x_{i}))}_{\ell_2}$ for a given input $x_i$. As mentioned earlier in the introduction section, domain-specific characteristics persist in both heads, which hinders the aggregation procedure of their outputs.

To address this issue, we expand the two heads to two branches, as illustrated in Fig. \ref{fig:struc}(b). The inference branch (\textcolor{orange}{orange} color) includes the inference head $F$ and the domain-specific head $F'$. Similarly, the adversarial branch (\textcolor{green}{green} color) contains the adversarial head $F_a$ and the adversarial domain-specific head $F'_a$. \ul{The purpose of $F'$ and $F'_a$ is to disentangle and extract domain-specific representations, denoted as $F'(G(x_i))$ and $F'_a(G(x_i))$.} Enlarging the discrepancy between these representations is beneficial for learning invariant knowledge \cite{hadsell2006dimensionality}.

By utilizing domain-specific representations, we obtain intermediate representations as $F(G(x_i))-F'(G(x_i))$ (the \textcolor{orange}{orange} square in Fig. \ref{fig:struc}b) and $F_a(G(x_i))-F'_a(G(x_i))$ (the \textcolor{green}{green} square in Fig. \ref{fig:struc}b). \ul{These representations serve as bridges between the two branches, and minimizing their discrepancy helps to reduce the difficulty of knowledge transfer.} In conclusion, the discrepancy measurement $\mathcal{L}_{dl}$ between the two branches contains the aggregation of intermediate representations $\mathcal{L}_{inter}$ and the segregation of specific representations $\mathcal{L}_{spec}$:
\begin{equation}
    \label{eq:dl}
    \mathcal{L}_{dl}(x_{i}) = \mathcal{L}_{inter}(x_{i}) - \mathcal{L}_{spec}(x_{i}),
\end{equation}
where the first term corresponds to the minimization of intermediate representations' discrepancy, and the second indicates the maximization of specific representations' discrepancy. Further details on how we compute these two terms separately are included in Sec. \ref{sec:dl}.

\subsection{Discrepancy Measurement}
\label{sec:dl}
In this section, we propose a new discrepancy measurement that is designed for the pose estimation problem, especially for the min-max game. Hypothesis disparity frameworks in domain adaptation are inspired by Maximum Classifier Discrepancy (MCD) \cite{saito2018maximum, xin2024vmt}, which maximizes and minimizes hypothesis discrepancy to form the min-max game. For some existing methods like \cite{jiang2021regressive,jin2022branch,han2022transpar}, MSE or KL divergence is applied as their discrepancy losses, but these losses are not enough for DA in HPE tasks. For HPE, the outputs of the regressors are much more complex than classification outputs.  The outputs of classification are just digits, but the regressors generate one heatmap for each keypoint, so simple MSE or KL divergence between heatmaps from different hypotheses cannot show the hypothesis disparity completely. Therefore, we introduce Maximum Mean Discrepancy (MMD) \cite{gretton2012kernel}, which estimates the distance between two hypotheses based on observed samples:
\begin{equation}
\label{eq:mmd}
    D_{\mathcal{H}}(p,q) = \mathcal{MMD}(h_{i},h_{j}) = \left\Vert \phi(h_{i}) - \phi(h_{j}) \right\Vert_{\mathcal{H}}^{2},
\end{equation}
where $h_{i}$ and $h_{j}$ are distinctive outputs and $\phi(\cdot)$ is a map that projects them to the Reproducing Kernel Hilbert Space (RKHS) and is related to a Gaussian kernel $\kappa$ as $\kappa(x_1,x_2)=\langle\phi(x_1), \phi(x_2)\rangle$, where $\langle\cdot,\cdot\rangle$ denotes the vector inner product. $p$ 
and $q$ are different hypotheses. We generalize MMD to the pose estimation, assuming $\boldsymbol{H_{i}}=F(G(x_{i}))= \begin{bmatrix}
        h_{i,1}, h_{i,2},\cdots,h_{i,K}
\end{bmatrix}^{\top}$, then MMD between two keypoints is defined as: 
\begin{equation}
\label{eq:mmd-keypoint}
    \mathcal{MMD}_{keypoint}(h_{i,m}, h_{j,n}) =   \left\Vert \phi(h_{i,m}) - \phi(h_{j,n}) \right\Vert_{\mathcal{H}}^{2},
\end{equation}where $i$ and $j$ represent different outputs, while $m$ and $n$ correspond to distinctive keypoints. 

\begin{figure}[t]
  \centering
  \includegraphics[width=1.0\linewidth]{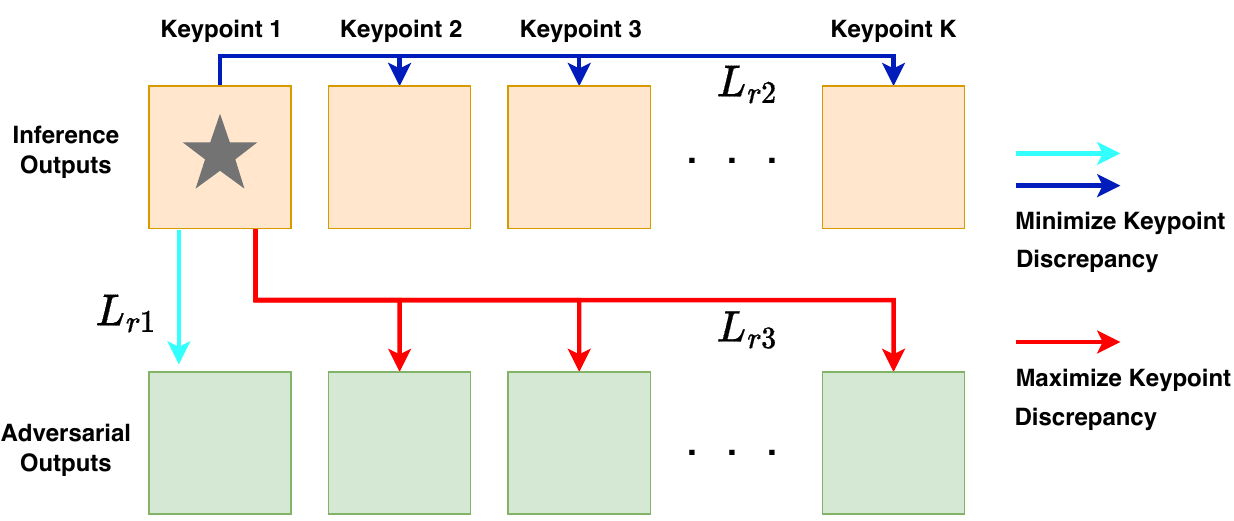}
   \caption{Proposed discrepancy loss. Here the keypoint with a star is used for illustration. Keypoints connected with a cyan arrow are inter-hypothesis identical ($r_1$ relation). $L_{r1}$ in Eq. \eqref{eq:r123} is used to describe their discrepancy. Keypoints marked with blue arrows are intra-hypothesis non-identical ($r_2$ relation) and represented with $L_{r2}$ in Eq. \eqref{eq:r123}. Keypoints linked with red arrows are inter-hypothesis non-identical ($r_3$ relation) computed by $L_{r3}$ in Eq. \eqref{eq:r123}.
   }%
   \vspace{-12pt}
   \label{fig:pose_align}
\end{figure}

\begin{figure*}[!ht]
  \centering
   \includegraphics[width=1.0\linewidth]{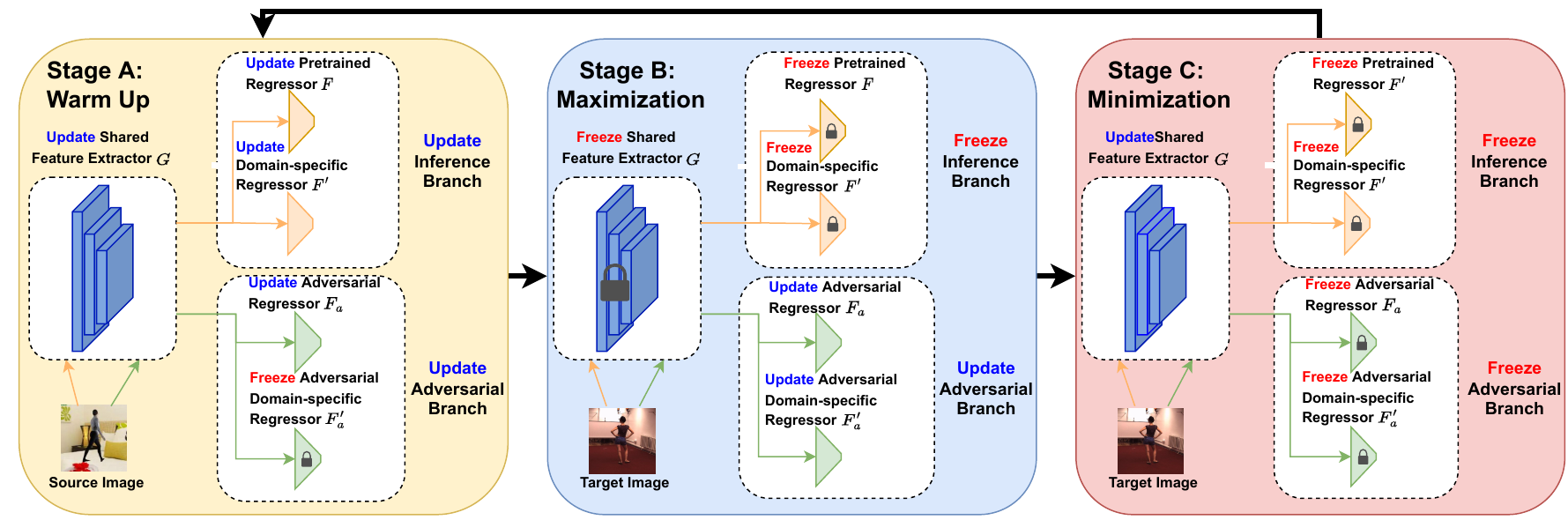}\vspace{-0mm}
   \caption{Overall framework of our method. It includes a shared feature extractor $G$ and two regression branches, the inference branch, and the adversarial branch. Each branch has two regressors, one for general-purpose estimation and the other for domain-specific estimation.
   }\vspace{-15pt}
   \label{fig:framework}
\end{figure*}

Moreover, the relations between the two hypothesis outputs are more complicated. For probabilistic outputs, we just deal with the digits at the same location one by one with $\ell_1$ distance. But for pose estimation, three types of relations matter as shown in Fig. \ref{fig:pose_align}. One is inter-hypothesis identical relation ($r_1$ relation). Similar to MCD-based methods \cite{jiang2021regressive,han2022transpar,jin2022branch}, we aim to minimize their discrepancy. Another relation is intra-hypothesis non-identical relation ($r_2$ relation). Since they are generated by identical regressors, it is expected that they keep closer, so we still minimize their hypothesis discrepancy. The last relation is the inter-hypothesis non-identical relation ($r_3$ relation).  Since they are generated by different regressors, we enlarge their hypothesis discrepancy. Based on this, we use $\mathcal{L}_{inter}$ to show the discrepancy between two intermediate outputs, $\boldsymbol{H^{*}}$ from inference branch and $\boldsymbol{H_{a}^{*}}$ from adversarial branch:  
\begin{equation}
    \label{eq:mmd-it}
    \mathcal{L}_{inter}(x_{i}^{t}) = \mathcal{MMD}_{output}(\boldsymbol{H^{*}}, \boldsymbol{H^{*}_{a}}) = L_{r1} + L_{r2} - L_{r3}, \\
\end{equation}
\begin{equation}
\label{eq:r123}
\left\{\begin{aligned}
    L_{r1} &= \frac{1}{K} \sum_{m=1}^{k} \mathcal{MMD}_{keypoint}(h_{m}^{*},h_{a,m}^{*})
    \\
    L_{r2} &= \frac{1}{K(K-1)} \sum_{m=1}^{k}  \sum_{n\ne m}  \mathcal{MMD}_{keypoint}(h_{m}^{*},h_{n}^{*})\\
    L_{r3} &= \frac{1}{K(K-1)} \sum_{m=1}^{k}  \sum_{n\ne m}  \mathcal{MMD}_{keypoint}(h_{m}^{*},h_{a,n}^{*})
\end{aligned} \right.
\end{equation}

Here $*$ means intermediate, while $m$ and $n$ correspond to different keypoints. $L_{r1}$ is for the inter-hypothesis identical keypoint discrepancy, while $L_{r2}$ is for the intra-hypothesis non-identical discrepancy, and $L_{r3}$ is for the inter-hypothesis non-identical situation. Similarly, we get the discrepancy between specific regressors' outputs as:
\begin{equation}
    \label{eq:mmd-sp}
    \mathcal{L}_{spec}(x_{i}^{t}) = \mathcal{MMD}_{output}(F'(G(x_{i}^{t})), F'_{a}(G(x_{i}^{t}))).
\end{equation}

Therefore, with Eq. \ref{eq:mmd-it} and Eq. \ref{eq:mmd-sp}, we can compute the proposed hypothesis discrepancy loss in Eq. \ref{eq:dl}.

\subsection{Overall Objectives and Optimizations}

In this part, we summarize one iteration of adversarial adaptation including three stages, as shown in Fig. \ref{fig:framework}. Stage A is for warm-up, Stage B tries to maximize branch discrepancy, and Stage C aims to minimize branch discrepancy.

\noindent\textbf{Stage A: Warm-up Stage.} At this stage, we use source data to warm up two untrained regressors $F'$ and $F_{a}$ at the beginning, and only $F'_{a}$ is fixed here. The reason is that we need to use the two regressors in each branch to represent intermediate domains. Since $F$ is well-pretrained, the difference between $F(G(\cdot))$ and $F'(G(\cdot))$ can be maintained. But if we warm up $F_{a}$ and $F'_{a}$ simultaneously with identical losses, $F_{a}(G(\cdot)) - F'_{a}(G(\cdot))$ will become very close to zero and cannot represent intermediate domain, hence harming the proceeding stages. 
Furthermore, this step is implemented to mitigate the issue of forgetting on source domain as the number of training epochs increases. The objective can be represented as: 
\begin{equation}
\begin{multlined}
 \label{eq:stage1}
\min_{G,F,F',F_{a}}\mathbb{E}_{x^{s}_{i},y^{s}_{i} \in \mathcal{S}}\mathcal{L}_{mse}(F(G(x_{i}^{s})), T^{-1}(y_{i}^{s})) \\
    + \alpha_{1} \mathcal{L}_{mse}(F(G(x_{i}^{s})),F'(G(x_{i}^{s}))) \\
    + \alpha_{2} \mathcal{L}_{mse}(F(G(x_{i}^{s})),F_{a}(G(x_{i}^{s}))).
\end{multlined}
\end{equation}

\noindent\textbf{Stage B: Maximization Stage.} At this stage, we start the \emph{max} process of the min-max game. In our proposed IDF, the goal is to maximize the discrepancy between the two branches, which is achieved by enlarging the discrepancy between intermediate representations and reducing the discrepancy between specific representations simultaneously. The shared feature extractor and inference branch are fixed, and only $F_{a}$ and $F'_{a}$ from the adversarial branch are updated:
\begin{equation}
 \label{eq:stage2}
    \max_{F_{a},F'_{a}}\mathbb{E}_{x^{t}_{i}\in \mathcal{T}}\mathcal{L}_{mse}(F(G(x_{i}^{t})), F_{a}(G(x_{i}^{t}))) + \beta \mathcal{L}_{dl}(x^{t}_{i}).
\end{equation} Without this stage, the outputs from two branches can be very similar and cannot detect target elements outside the support of the source. 

\noindent\textbf{Stage C: Minimization Stage.} For the final stage, we tackle the \emph{min} process of the min-max game. We hope that the discrepancy between the two branches can be minimized, which is achieved by reducing the discrepancy between intermediate representations and enlarging the discrepancy between specific representations at the same time. In such a case, only $G$ is updated and all the other parts are fixed. The shared feature extractor confuses the two branches and makes them closer:
\begin{equation}
\begin{multlined}
 \label{eq:stage3}
    \min_{G} \mathbb{E}_{x^{t}_{i}\in \mathcal{T}}\mathcal{L}_{mse}(F(G(x_{i}^{t})), F_{a}(G(x_{i}^{t})))\\
    + \mathcal{L}_{oks} (T(F(G(x_{i}^{t}))), T(F_{a}(G(x_{i}^{t})))) \\ 
    + \gamma \mathcal{L}_{dl}(x^{t}_{i}).  
\end{multlined}
\end{equation}

We still use $\mathcal{L}_{mse}$ and $\mathcal{L}_{dl}$ at this stage but optimize a different objective. This stage encourages target information to be detected with the source support. To address similar concerns about the heatmap errors during the transformation as in the pretraining stage, we add $\mathcal{L}_{oks}$. We provide PyTorch-like pseudo-code of our method in \textcolor{blue}{Supplementary}.

\section{Experiments}

\subsection{Datasets}

We use three human-pose datasets and three hand-pose datasets to validate our approach. \textbf{SURREAL} \cite{varol2017learning} is the source human pose dataset and offers six million synthetic human pose images. \textbf{Human3.6M} \cite{ionescu2013human3} is one of the target human pose datasets and is a widely used real-world dataset in the community. It contains 3.6 million images split into 7 folds, and we treat S1, S5, S6, S7, S8 as the training set and S9, S11 as the testing set according to \cite{jiang2021regressive} and \cite{kim2022unified}. \textbf{Leeds Sports Pose} \cite{johnson2010clustered} (LSP) is the other target human dataset. It is a real-world dataset with 2,000 images and all of them are used for adaptation. Human pose tasks focus on two domain adaptation tasks SURREAL $\rightarrow$ Human3.6M and SURREAL $\rightarrow$ LSP. \textbf{Rendered Hand Pose Dataset} \cite{zimmermann2017learning} (RHD) is the source hand dataset with 43,986 synthetic hand images. 41,258 of them are used for training and the rest 2,728 are tackled with validation. \textbf{Hand-3D-Studio} \cite{zhao2020h3d} (H3D) is one of the two target hand datasets, which includes 22,000 real-world frames. Following the policy from \cite{jiang2021regressive} and \cite{kim2022unified}, 18,800 frames are used for training and the others are for testing. \textbf{FreiHand} \cite{zimmermann2019freihand} is the other target hand dataset, providing 130k real-world images, and we use all of them for our adaptation task. Hand pose tasks focus on two domain adaptation tasks RHD $\rightarrow$ H3D and RHD $\rightarrow$ FreiHand.

\subsection{Implementation Details} 

Following previous works \cite{jiang2021regressive,kim2022unified}, Simple Baseline \cite{xiao2018simple} with ResNet-101 \cite{he2016deep} is adopted as the HPE backbone. In the \emph{pretrain} process, 70 epochs are conducted, each with 500 iterations. The initial pretrain learning rate is 1e-3 with Adam \cite{kingma2014adam} optimizer. The learning rate drops to 1e-4 at 30 epochs and 1e-5 at 50 epochs. As for the \emph{adaptation} process, 40 epochs are executed with a total of 30,000 iterations. SGD \cite{amari1993backpropagation} is applied as the optimizer with momentum 0.9 and weight decay 1e-4. The initial learning rate of the shared feature extractor is 1e-3, and that of each regressor is 1e-2. We use the learning rate scheduler with the same strategy used by \cite{jiang2021regressive}. 

Regarding hyperparameters, we set $\alpha_{1} = \alpha_{2} = 0.5$, and select $\beta$ as $0.2$. Besides, $\gamma$ is set to be $0.55$. 
To represent the maximization process of Stage B better, we use the same strategy as \cite{jiang2021regressive} to minimize negative heatmaps built from spatial probability. All experiments are conducted on Nvidia RTX A5000 GPUs. 

\subsection{Main Results }

\textbf{Baselines.} Since this paper focuses on \emph{domain adaptive 2D pose estimation}, four recent methods \textbf{CC-SSL} \cite{mu2020learning}, \textbf{MDAM} \cite{li2021synthetic}, \textbf{RegDA} \cite{jiang2021regressive} and \textbf{UniFrame} \cite{kim2022unified} with SOTA performances are chosen as baselines for comparison. Besides, we add \textbf{Source-only} as another baseline for all four tasks to evaluate the effectiveness of adaptation. Since RHD $\rightarrow$ H3D and SURREAL $\rightarrow$ Human3.6M use separate sets for the target domain during training and validation, we add \textbf{Oracle} as another baseline, which is jointly trained with target annotations.  

\noindent\textbf{Metrics.} Percentage of Correct Keypoint (PCK) is adopted as in previous work. We set the ratio of correct prediction as $5\%$ and report PCK@0.05 in Tables \ref{tab:rhd2h3d}-\ref{tab:surreal2lsp}. Apart from overall keypoint accuracy, joints are divided into several part segments and measured by specific metrics following \cite{jiang2021regressive,kim2022unified}. For the 21-keypoint hand skeleton, MCP, PIP, DIP, and Fin are applied. For the 18-keypoint human skeleton, Sld, Elb, Wrist, Hip, Knee, and Ankle are selected. 

\noindent\textbf{Quantitative Results.} Table \ref{tab:rhd2h3d} and Table \ref{tab:rhd2fre} are for two hand tasks RHD$\rightarrow$H3D and RHD$\rightarrow$FreiHand. Our method reaches state-of-the-art for both tasks and achieves a significant lead of $3.4\%$ and $2.8\%$. Table \ref{tab:surreal2h36m} and Table \ref{tab:surreal2lsp} are for human tasks SURREAL$\rightarrow$Human3.6M and SURREAL$\rightarrow$LSP. Our model again outperforms all the baselines (except the Oracle) by a considerable margin of $3.2\%$ and $2.8\%$, respectively. 

\begin{table}[!ht]
    \scriptsize
    \centering
    \caption{PCK@0.05 on RHD $\rightarrow$ H3D Task}\vspace{-5pt}
    \resizebox{1.0\linewidth}{!}{%
    \begin{tabular}{ccccccccc}
          \toprule
          Method &  MCP &  PIP  &  DIP & Fin & All \\
         \hline
         {Source-only} & 67.4 & 64.2 & 63.3 & 54.8 & 61.8\\
         {Oracle} & 97.7 & 97.2 & 95.7 & 92.5 & 95.8\\
         \hline
         CC-SSL \cite{mu2020learning}  & 81.5 & 79.9 & 74.4 & 64.0 & 75.1 \\
         MDAM \cite{li2021synthetic} & 82.3 & 79.6 & 72.3 & 61.5 & 74.1\\
         RegDA \cite{jiang2021regressive}  & 79.6 & 74.4 & 71.2 & 62.9 & 72.5\\
         UniFrame \cite{kim2022unified}  & 86.7 & 84.6 & 78.9 & 68.1 & 79.6\\
         \hline
         {Ours} &   \textbf{88.9} &  \textbf{88.1} &  \textbf{82.1} &  \textbf{71.9} &  \textbf{83.0}\\
         \bottomrule
    \end{tabular}%
    }
\label{tab:rhd2h3d}
\end{table}

\begin{table}[!ht]
    \scriptsize
    \centering
    \caption{PCK@0.05 on RHD $\rightarrow$ FreiHand Task}\vspace{-5pt}
    \resizebox{0.8\linewidth}{!}{%
    \begin{tabular}{ccccccccc}
          \toprule
          Method &  MCP &  PIP  &  DIP & Fin & All \\
         \hline
         {Source-only} & 35.2 & 50.1 & 54.8 & 50.7 & 46.8\\
         \hline
         CC-SSL \cite{mu2020learning}  & 37.4 & 48.2 & 50.1 & 46.5 & 43.8\\
         MDAM \cite{li2021synthetic}  & 32.3 & 48.1 & 51.7 & 47.3 & 45.1\\
         RegDA \cite{jiang2021regressive}  & 40.9 & 55.0 & 58.2 & 53.1 & 51.1\\
         UniFrame \cite{kim2022unified}  & 43.5 & 64.0 & 67.4 & 62.4 & 58.5\\
         \hline
         {Ours} &  \textbf{46.0} & \textbf{65.2} & \textbf{69.5} & \textbf{63.7} & \textbf{61.3}\\    
         \bottomrule
    \end{tabular}%
    }
    \vspace{-5pt}
\label{tab:rhd2fre}
\end{table}

\begin{table}[!ht]
    \scriptsize
    \centering
    \caption{PCK@0.05 on SURREAL $\rightarrow$ Human3.6M Task}\vspace{-5pt}
    \resizebox{1.0\linewidth}{!}{%
    \begin{tabular}{ccccccccc}
          \toprule
          Method &  Sld & Elb & Wrist & Hip & Knee & Ankle & All \\
         \hline
         {Source-only} & 69.4 & 75.4 & 66.4 & 37.9 & 77.3 & 77.7 & 67.3\\
         {Oracle} & 95.3 & 91.8 & 86.9 & 95.6 & 94.1 & 93.6 & 92.9\\
         \hline
         CC-SSL \cite{mu2020learning} & 44.3 & 68.5 & 55.2 & 22.2 & 62.3 & 57.8 & 51.7 \\
         MDAM \cite{li2021synthetic} & 51.7 & 83.1 & 68.9 & 17.7 & 79.4 & 76.6 & 62.9\\
         RegDA \cite{jiang2021regressive} & 73.3 & 86.4 & 72.8 & 54.8 & 82.0 & 84.4 & 75.6\\
         UniFrame \cite{kim2022unified} & 78.1 & 89.6 & 81.1 & 52.6 & 85.3 & 87.1 & 79.0\\
         \hline
         {Ours} & \textbf{80.8} & \textbf{91.6} & \textbf{82.3} & \textbf{56.0} & \textbf{87.8} & \textbf{88.9} & \textbf{82.2}\\    
         \bottomrule
    \end{tabular}%
    }
    \vspace{-5pt}
\label{tab:surreal2h36m}
\end{table}

\begin{table}[!ht]
    \scriptsize
    \centering
    \caption{PCK@0.05 on SURREAL $\rightarrow$ LSP  Task}      \vspace{-5pt}
    \resizebox{1.0\linewidth}{!}{%
    \begin{tabular}{ccccccccc}
          \toprule
          Method &  Sld & Elb & Wrist & Hip & Knee & Ankle & All \\
         \hline
         {Source-only} & 51.5 & 65.0 & 62.9 & 68.0 & 68.7 & 67.4 & 63.9 \\
         \hline
         CC-SSL \cite{mu2020learning} & 36.8 & 66.3 & 63.9 & 59.6 & 67.3 & 70.4 & 60.7 \\
         MDAM \cite{li2021synthetic} & 61.4 & 77.7 & 75.5 & 65.8 & 76.7 & 78.3 & 69.2 \\
         RegDA \cite{jiang2021regressive} & 62.7 & 76.7 & 71.1 & 81.0 & 80.3 & 75.3 & 74.6\\
         UniFrame \cite{kim2022unified} & 69.2 & 84.9 & 83.3 & 85.5 & 84.7 & 84.3 & 82.0\\
         \hline
         {Ours} & \textbf{72.1} & \textbf{86.4} & \textbf{85.2} & \textbf{87.7} & \textbf{87.0} & \textbf{86.2} & \textbf{84.8}\\    
         \bottomrule
    \end{tabular}%
    }
    \vspace{-10pt}
\label{tab:surreal2lsp}
\end{table}

\begin{figure}[t]
  \centering
   \includegraphics[width=1.00\linewidth]{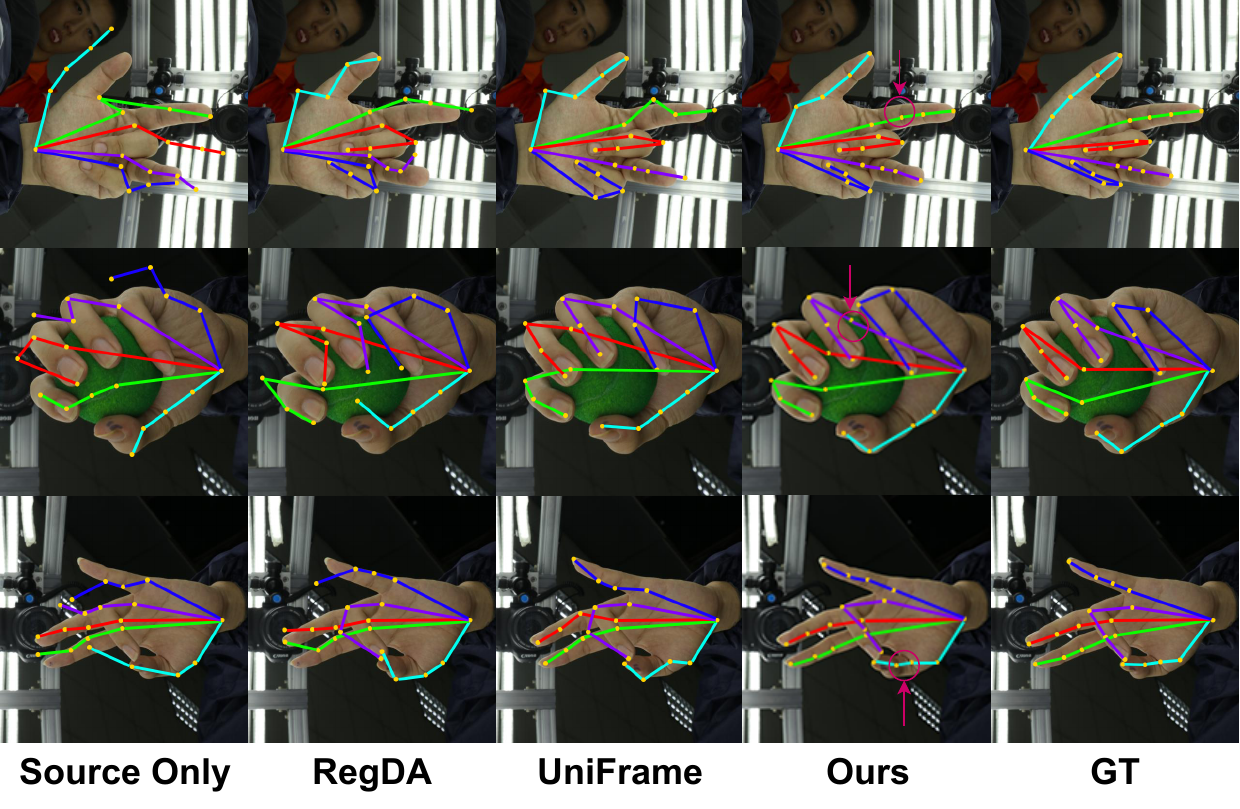}

   \caption{ Qualitative results on H3D dataset 
   } %
   \vspace{-10pt}
   \label{fig:h3d}
\end{figure}

\begin{figure}[t]
  \centering
   \includegraphics[width=1.00\linewidth]{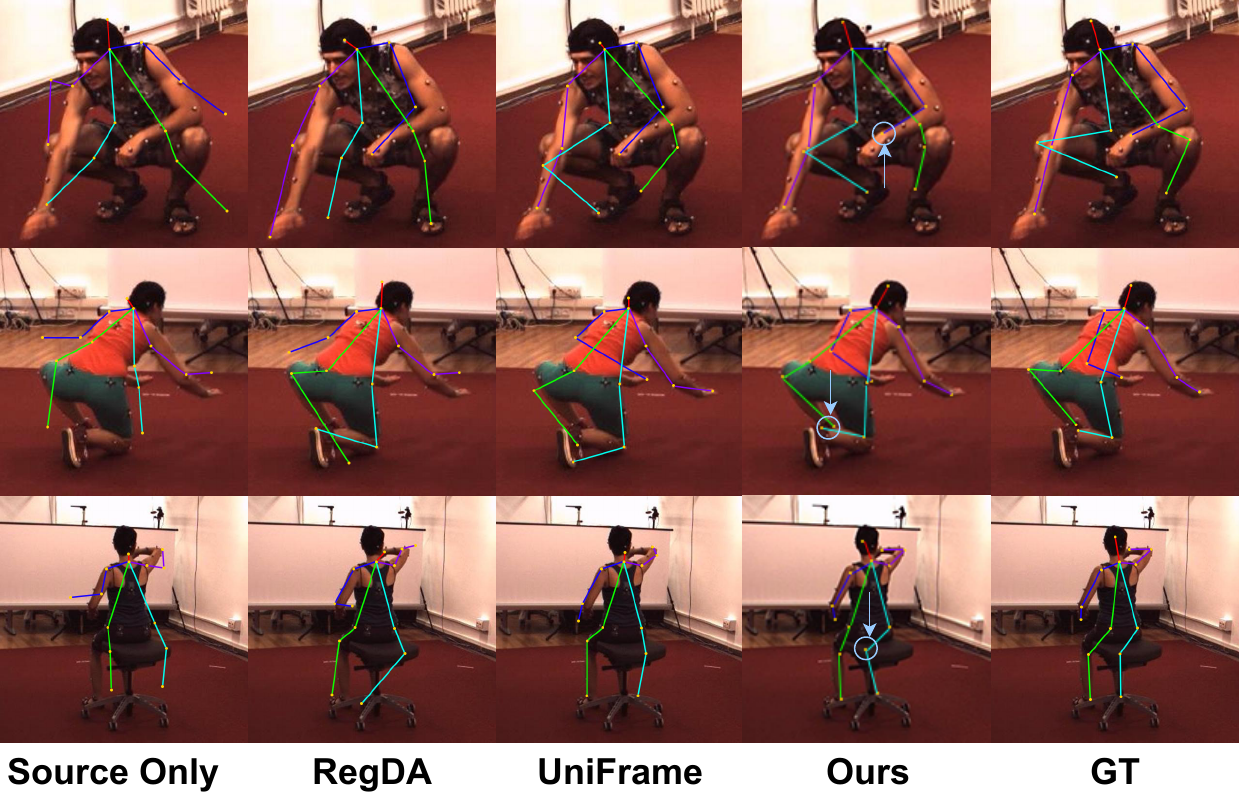}

   \caption{ Qualitative results on Human3.6M dataset
   }\vspace{-10pt}
   \label{fig:h36m}
\end{figure}

\begin{figure}[!ht]
  \centering
   \includegraphics[width=1.0\linewidth]{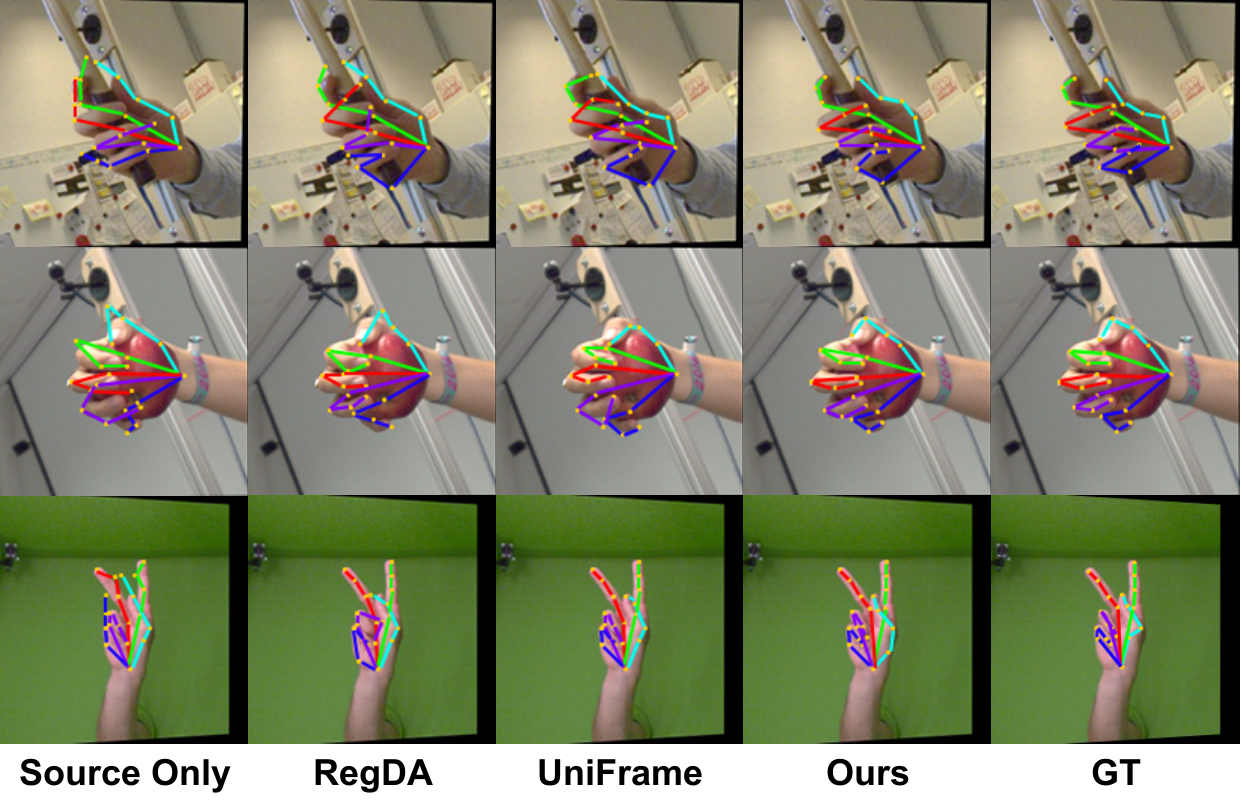}

   \caption{Qualitative results on FreiHand dataset}\vspace{-10pt}
   \label{fig:freihand}
\end{figure}

\begin{figure}[!ht]
  \centering
   \includegraphics[width=1.0\linewidth]{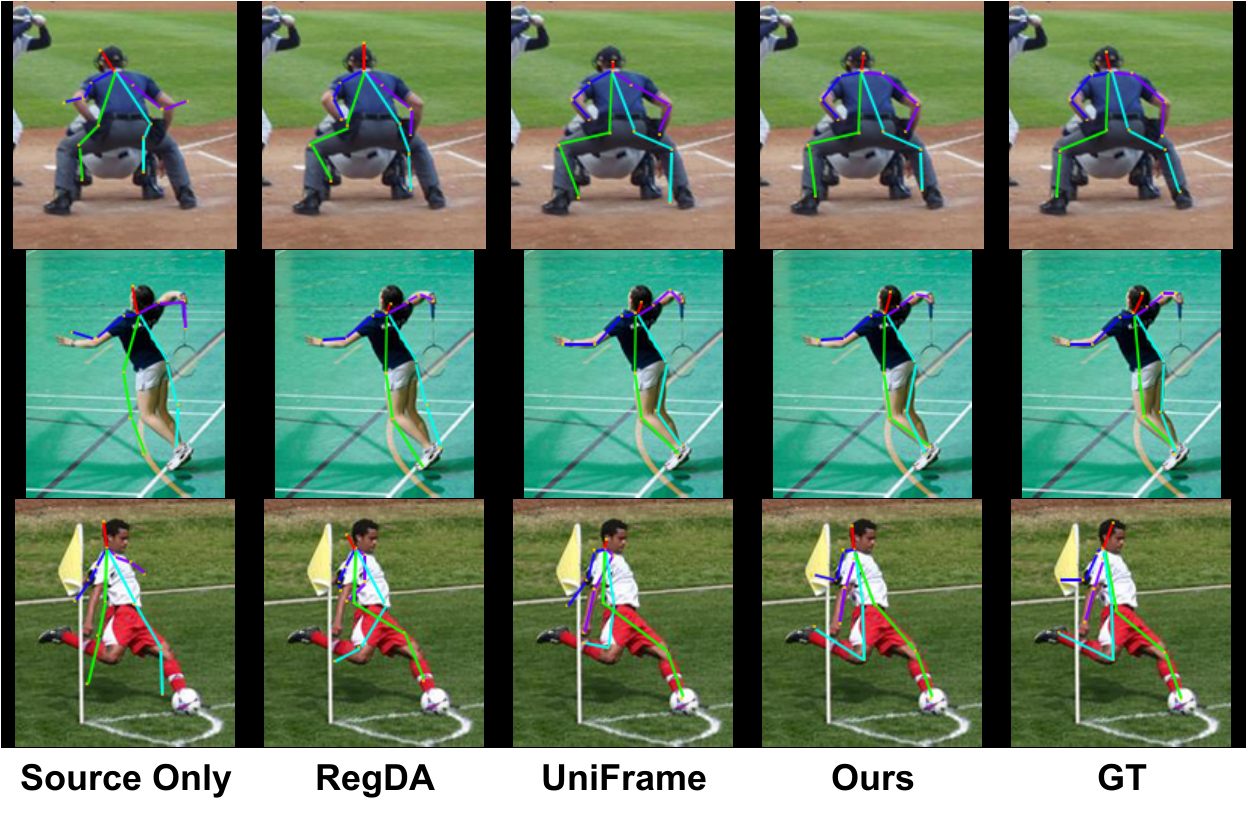}

   \caption{ Qualitative results on LSP dataset
   }\vspace{-10pt}
   \label{fig:lsp}
\end{figure}

\noindent\textbf{Qualitative Results.} Figs. \ref{fig:h3d} and \ref{fig:h36m} show qualitative results on H3D and Human3.6M, while Figs. \ref{fig:freihand} and \ref{fig:lsp} show qualitative results on FreiHand and LSP. We use Source only, RegDA \cite{jiang2021regressive}, UniFrame \cite{kim2022unified}, Ours, and Ground Truth (GT) for qualitative comparison. It is evident from these figures that our method provides better adaptation outcomes with more accurate human or hand poses. 

\subsection{Ablation Study on Discrepancy Loss}
We further conduct a detailed ablation study on the three terms in Eq. \ref{eq:r123} using two tasks RHD$\rightarrow$H3D and SURREAL$\rightarrow$Human3.6M. According to the three relations described in Section \ref{sec:dl}, seven combinations $r_1$, $r_2$, $r_3$, $r_1$ \& $r_2$,  $r_1$ \& $r_3$, $r_2$ \& $r_3$, and $r_1$ \& $r_2$ \& $r_3$ are implemented and the results are listed in Tables \ref{tab:rh-ab-dl} and \ref{tab:sh-ab-dl}. 

\begin{table}[!ht]
    \scriptsize
    \centering
    \caption{Ablation of discrepancy loss terms on RHD $\rightarrow$ H3D task}
    \resizebox{1.0\linewidth}{!}{%
    \begin{tabular}{ccccccccc}
          \toprule
          Method &  MCP &  PIP  &  DIP & Fin & All \\
         \hline
         {$r_1$} & 85.5 & 86.2 & 80.7 & 72.0 & 79.4\\
         {$r_2$} & 82.1 & 81.7 & 76.9 & 70.1 & 74.3\\
         {$r_3$} & 83.7 & 82.6 & 77.7 & 69.5 & 74.0\\
         {$r_1$ \& $r_2$} & 88.8 & 88.0 & 81.6 & 72.5 & 82.2\\
         {$r_1 $ \& $r_3$} & 87.4 & 87.3 & 81.1 & 72.0 & 80.6\\
         {$r_2 $ \& $r_3$} & 88.0 & 87.7 & 80.6 & 71.9 & 80.8\\
         {$r_1$ \& $r_2$ \& $r_3$} & 89.6 & 88.5 & 82.4 & 73.0 & 83.5\\    
         \bottomrule
    \end{tabular}%
    }
    \vspace{-10pt}
\label{tab:rh-ab-dl}
\end{table}

\begin{table}[!ht]
    \scriptsize
    \centering
    \caption{Ablation of discrepancy loss terms on SURREAL $\rightarrow$ Human3.6M task}
    \resizebox{1.0\linewidth}{!}{%
    \begin{tabular}{ccccccccc}
          \toprule
          Method &  Sld & Elb & Wrist & Hip & Knee & Ankle & All \\
         \hline
         {$r_1$} & 77.3 & 91.0 & 80.4 & 54.5 & 84.0 & 85.8 & 78.6\\
         {$r_2$} & 75.7 & 89.4 & 79.2 & 53.0 & 82.6 & 83.9 & 76.8\\
         {$r_3$} & 76.1 & 89.8 & 78.2 & 52.5 & 82.5 & 84.3 & 77.3\\
         {$r_1$ \& $r_2$} & 80.6 & 91.9 & 81.5 & 55.7 & 85.9 & 87.4 & 81.6\\
         {$r_1$ \& $r_3$} & 78.9 & 91.6 & 80.3 & 55.1 & 83.3 & 87.0 & 80.4\\
         {$r_2$ \& $r_3$} & 78.0 & 90.9 & 81.4 & 54.9 & 84.3 & 86.6 & 79.8\\
         {$r_1$ \& $r_2$ \& $r_3$} &  81.5 & 92.3 & 82.9 & 56.4 & 88.8 & 89.2 & 82.7\\    
         \bottomrule
    \end{tabular}%
    }
    \vspace{-8pt}
\label{tab:sh-ab-dl}
\end{table}

We observe that \emph{all these three relations have an impact on the effectiveness of our proposed discrepancy loss.} Simply considering one relation is not ideal, because even the best situation $r_1$ still underperforms the baseline UniFrame. On the other hand, \emph{the removal of $r_1$, $r_2$ or $r_3$ will degrade the model's performance.} Simply removing $r_1$ leads to a decrease of $2.7\%$ in RHD$\rightarrow$H3D and $2.9\%$ in SURREAL$\rightarrow$Human3.6M. Simply removing $r_2$ leads to a decrease of $2.9\%$ in RHD$\rightarrow$H3D and $2.3\%$ in SURREAL$\rightarrow$Human3.6M. As for $r_3$, eliminating it causes a drop of $1.3\%$ in RHD$\rightarrow$H3D and $1.1\%$ in SURREAL$\rightarrow$Human3.6M. More results on other tasks are provided in \textcolor{blue}{Supplementary}.

\subsection{Ablation Study on Network Structures}
\label{sec:compare}

\subsubsection{Network Structures Comparison} 
\begin{figure}[!h]
\vspace{-5pt}
  \centering
   \includegraphics[width=0.95\linewidth]{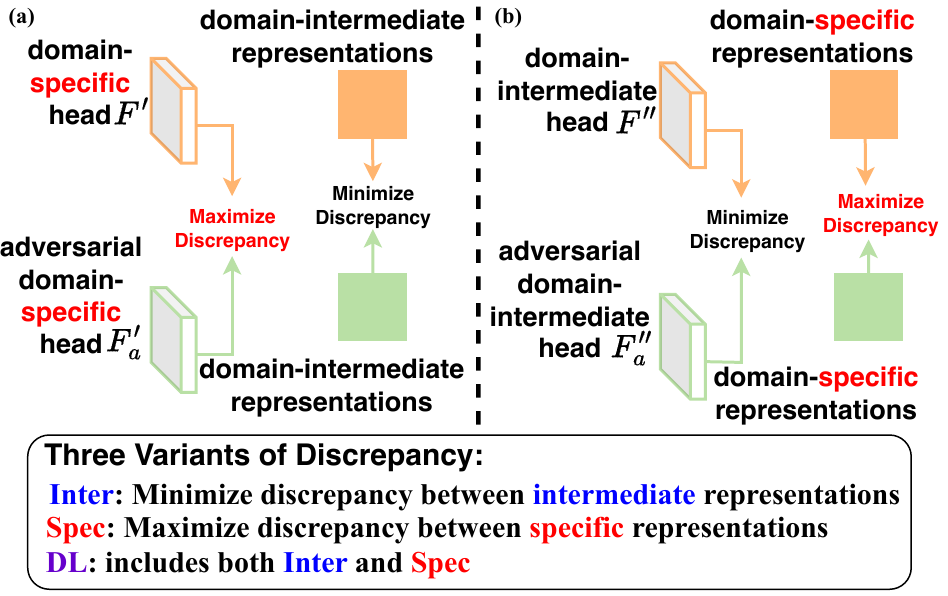}

   \caption{\small{Comparisons between (a) our proposed Intermediate Domain Framework (IDF), and (b) Alternative Intermediate Domain Framework (AIDF). Detailed illustrations are in \textbf{Sec. \ref{sec:compare}}.} }%
   \vspace{-10pt}
   \label{fig:alt}
\end{figure}

To validate the effectiveness of our proposed Intermediate Domain Framework (\textbf{IDF}), we compare it with the {\textbf{Baseline}} and an Alternative Intermediate Domain Framework ({\textbf{AIDF}}). 
Specifically, `Baseline' is the conventional structure as shown in Fig. \ref{fig:struc} (a).
Since there exists only one head in each branch of the Baseline, the domain-specific characteristics persist in both heads, which hinders the aggregation procedure of their outputs. Our proposed `IDF' (Fig. \ref{fig:struc} (b)) solves this issue by adding two \ul{explicit specific heads} to two branches separately, obtaining two intermediate representations for segregating specific features and aggregate invariant features. A simplified graphical illustration of `IDF' is also presented in Fig. \ref{fig:alt} (a). Unlike our `IDF', `AIDF' utilizes two \ul{explicit intermediate heads} and obtains two specific representations indirectly.

\subsubsection{Variants of Discrepancy} 

Within each framework, we identify three discrepancy variants that govern the aggregation or segregation of representations from two distinct branches, as depicted in Fig. \ref{fig:alt}. The variant \textcolor{blue}{\textbf{Inter}} aims to minimize the discrepancy of intermediate representations, aligning with the \emph{aggregation} process. The variant \textcolor{red}{\textbf{Spec}} seeks to maximize the discrepancy between specific representations, aligning with the \emph{segregation} process. \textcolor{amethyst}{\textbf{DL}} is our proposed discrepancy measure formulated in Eq. \ref{eq:dl}, which encompasses both Inter and Spec aspects. The influence of these discrepancy variants on each framework is assessed and documented in Tables \ref{tab:rh-ab} and \ref{tab:sh-ab}.

Specifically, the terms `AIDF' and `IDF' in these tables refer to the utilization of only the inference and adversarial heads, respectively, for each branch (which is similar to the Baseline approach). The term `Baseline w/ DL' pertains to the reduction of discrepancy between the two heads of the Baseline, as it does not allow for the disentanglement of invariant and specific features.

\subsubsection{Observations} 
Based on the results in these two tables, we conclude several observations:
\begin{itemize}
    \item \emph{Adversarial models fail to exhibit enhanced performance by employing `IDF' or `AIDF' exclusively, without integrating any discrepancy variants.} However, when `IDF' or `AIDF' is combined with these variants, there is a significant performance boost. This occurs because when only two heads are used as in the `Baseline', the network structures do not effectively perform feature disentanglement, rendering them identical to the `Baseline'.
    \item \emph{Disentangling invariant and specific representations is necessary, and our `IDF' is better than `AIDF' in disentanglement.} Notably, in the category of methods using `w/ DL', `IDF w/ DL' achieves a performance gain of $3.7\%$ compared to `Baseline w/ DL' on SURREAL $\rightarrow$ Human3.6M, and it also outperforms `AIDF w/ DL' by $2.9\%$. The substantial improvements achieved by `AIDF' over the Baseline underscore the critical importance of disentangling invariant and specific representations within heads for improved adaptation. And the noticeable disparity between IDF's performance and that of `AIDF' suggests that the construction of explicit specific heads is a more effective approach compared to AIDF's method of employing explicit intermediate heads. This distinction arises from the neural network's inherent capability to more easily extract specific features as opposed to invariant features.
    \item \emph{Solely adopting `Inter' or `Spec' results in limited performance improvement. However, when these two components cooperate and form `DL', the performances show significant improvement.} Specifically, on the RHD $\rightarrow$ H3D task, applying only `Inter' to `IDF' leads to an increase of $2.3\%$, and using only `IDF w/ Spec' results in an improvement of $1.7\%$. In contrast, when `DL' is formed, the performance boost becomes substantial at $8.2\%$. The results show that the process of learning invariant representations relies on both aggregating domain-invariant features and segregating domain-specific features, and these two processes can help each other get better. 
\end{itemize} 

\vspace{-5pt}
\begin{table}[!ht]
    \scriptsize
    \centering
    \caption{Ablation study on RHD $\rightarrow$ H3D task}
    \vspace{-8pt}
    \resizebox{0.9\linewidth}{!}{%
    \begin{tabular}{ccccccccc}
          \toprule
          Method &  MCP &  PIP  &  DIP & Fin & All \\
         \hline
         {Baseline} & 82.2 & 75.9 & 72.6 & 63.0 & 74.5\\
         {Baseline w/ DL} & 83.0 & 77.6 & 74.7 & 65.5 & 75.9\\
         \hline

         {AIDF} & 82.2 & 76.1 & 73.0 & 63.5 & 74.6\\
         {AIDF w/ Inter} & 83.3 & 76.9 & 73.7 & 64.2 & 75.4\\
         {AIDF w/ Spec} & 82.6 & 76.5 & 73.6 & 63.8 & 74.9\\
         {AIDF w/ DL} & 84.7 & 79.9 & 75.3 & 67.0 & 77.8\\

         \hline
         {IDF} & 82.3 & 76.1 & 73.4 & 64.2 & 74.8\\
         {IDF w/ Inter} & 85.5 & 78.3 & 75.2 & 66.1 & 77.1\\
         {IDF w/ Spec} & 84.9 & 78.1 & 74.6 & 65.8 & 76.5\\
         {IDF w/ DL (Ours)} &  \textbf{88.9} &  \textbf{88.1} &  \textbf{82.1} &  \textbf{71.9} &  \textbf{83.0}\\
         
         \bottomrule
    \end{tabular}%
    }
   \vspace{-8pt}
\label{tab:rh-ab}
\end{table}

\vspace{-10pt}
\begin{table}[!ht]
    \scriptsize
    \centering
    \small
    \caption{Ablation study on SURREAL $\rightarrow$ Human3.6M}
    \vspace{-8pt}
    \resizebox{1.0\linewidth}{!}{%
    \begin{tabular}{ccccccccc}
          \toprule
          Method &  Sld & Elb & Wrist & Hip & Knee & Ankle & All \\
         \hline
         {Baseline} & 74.5 & 88.0 & 72.4 & 54.6 & 84.0 & 85.0 & 77.7\\
         {Baseline w/ DL} & 76.0 & 90.8 & 75.1 & 55.2 & 84.4 & 86.4 & 78.5\\
         \hline

         {AIDF} & 74.7 & 88.8 & 74.7 & 54.5 & 84.2 & 85.3 & 77.7\\
         {AIDF w/ Inter} & 75.6 & 89.3 & 75.4 & 55.0 & 84.6 & 86.0 & 78.8 \\
         {AIDF w/ Spec} & 75.3 & 88.8 & 75.3 & 54.7 & 84.0 & 85.9 & 78.1\\
         {AIDF w/ DL}  & 78.4 & 91.1 & 76.5 & 55.2 & 85.9 & 86.6 & 79.3\\

         \hline
         {IDF} & 74.7 & 89.5 & 75.0 & 54.8 & 84.3 & 85.7 & 77.8\\
         {IDF w/ Inter} & 76.4 & 91.0 & 77.9 & 55.3 & 85.6 & 87.2 & 80.7\\
         {IDF w/ Spec} & 75.5 & 90.7 & 77.3 & 55.0 & 86.2 & 86.8 & 80.3\\
         {IDF w/ DL (Ours)} & \textbf{80.8} & \textbf{91.6} & \textbf{82.3} & \textbf{56.0} & \textbf{87.8} & \textbf{88.9} & \textbf{82.2} \\
         \bottomrule
    \end{tabular}%
    }
    \vspace{-15pt}
\label{tab:sh-ab}
\end{table}

\subsection{Analysis of Training Efficiency} 

To assess the training efficiency in a fair manner, we conducted experiments using the RHD$\rightarrow$H3D task on a single Nvidia RTX A5000 GPU. By comparing the training times of RegDA, UniFrame, and our method, as presented in Table \ref{tab:time}, we observe that ours takes similar time as RegDA (while achieving significant improvements over RegDA), but considerably less time than UniFrame. This discrepancy in training times underscores the efficacy of our method in achieving superior performance while maintaining competitive training durations. Specifically, our method exhibits a balance between training efficiency and performance gains, making it a compelling choice for real-world applications where computational resources and time constraints are critical factors to consider. \emph{Moreover, the notable reduction in training time compared to UniFrame highlights the potential scalability and feasibility of our approach for large-scale deployment and practical usage scenarios.} 
\vspace{-10pt}
\begin{table}[!ht]
\centering
\caption{Comparisons of Training Time}
\vspace{-8pt}
\resizebox{1.0\linewidth}{!}{%
\begin{tabular}{c|ccc}
\toprule
 & RegDA & UniFrame & Ours \\
\hline
Style Transfer Time & 0 & 7h45min & 0 \\
Pretrain Epoch/Time & 70/3h05min & 70/3h10min & 70/3h07min \\
Adaptation Epoch/Time & 60/4h12min & 60/6h10min & 60/5h23min \\
All Time & \textbf{7h17min} & 17h05min & 8h29min \\
Performance (PCK@0.05) & 72.5 & 79.6 & \textbf{83.0} \\
\bottomrule
\end{tabular}}
\label{tab:time}
\end{table}
\vspace{-15pt}

\vspace{-2pt}
\section{Conclusion}
In this paper, we propose a novel domain adaptive pose estimation model, where we aim to alleviate the transfer difficulty via disentangling features into domain-intermediate and domain-specific components, along with aggregation and segregation of them separately. Moreover, we design a new discrepancy measurement that exploits various keypoint relationships and apply separate aggregation or segregation strategies to them. Extensive experiments are conducted on hand and human pose datasets, showing that our approach outperforms the state-of-the-art methods by significant margins.

{\small
\bibliographystyle{ieee}
\bibliography{egbib}

\begin{thebibliography}{10}\itemsep=-1pt

\bibitem{amari1993backpropagation}
S.-i. Amari.
\newblock Backpropagation and stochastic gradient descent method.
\newblock {\em Neurocomputing}, 5(4-5):185--196, 1993.

\bibitem{ben2010theory}
S.~Ben-David, J.~Blitzer, K.~Crammer, A.~Kulesza, F.~Pereira, and J.~W. Vaughan.
\newblock A theory of learning from different domains.
\newblock {\em Machine learning}, 79:151--175, 2010.

\bibitem{chen2018cascaded}
Y.~Chen, Z.~Wang, Y.~Peng, Z.~Zhang, G.~Yu, and J.~Sun.
\newblock Cascaded pyramid network for multi-person pose estimation.
\newblock In {\em Proceedings of the IEEE Conference on Computer Vision and Pattern Recognition}, pages 7103--7112, Salt Lake City, Utah, 2018. IEEE.

\bibitem{deng2009imagenet}
J.~Deng, W.~Dong, R.~Socher, L.-J. Li, K.~Li, and L.~Fei-Fei.
\newblock Imagenet: A large-scale hierarchical image database.
\newblock In {\em 2009 IEEE conference on computer vision and pattern recognition}, pages 248--255, Miami, Florida, 2009. IEEE.

\bibitem{deng2021cluster}
W.~Deng, Q.~Liao, L.~Zhao, D.~Guo, G.~Kuang, D.~Hu, and L.~Liu.
\newblock Joint clustering and discriminative feature alignment for unsupervised domain adaptation.
\newblock {\em IEEE Transactions on Image Processing: a Publication of the IEEE Signal Processing Society}, 30:7842--7855, 2021.

\bibitem{dosovitskiy2020image}
A.~Dosovitskiy, L.~Beyer, A.~Kolesnikov, D.~Weissenborn, X.~Zhai, T.~Unterthiner, M.~Dehghani, M.~Minderer, G.~Heigold, S.~Gelly, J.~Uszkoreit, and N.~Houlsby.
\newblock An image is worth 16x16 words: Transformers for image recognition at scale.
\newblock In {\em International Conference on Learning Representations}, page N/A, Vienna,Austria, 2021. Open Review.

\bibitem{ganin2015unsupervised}
Y.~Ganin and V.~Lempitsky.
\newblock Unsupervised domain adaptation by backpropagation.
\newblock In {\em International Conference on Machine Learning}, pages 1180--1189, Lille, France, 2015. PMLR.

\bibitem{goodfellow2020generative}
I.~Goodfellow, J.~Pouget-Abadie, M.~Mirza, B.~Xu, D.~Warde-Farley, S.~Ozair, A.~Courville, and Y.~Bengio.
\newblock Generative adversarial networks.
\newblock {\em Communications of the ACM}, 63(11):139--144, 2020.

\bibitem{gretton2012kernel}
A.~Gretton, K.~M. Borgwardt, M.~J. Rasch, B.~Sch{\"o}lkopf, and A.~Smola.
\newblock A kernel two-sample test.
\newblock {\em The Journal of Machine Learning Research}, 13(1):723--773, 2012.

\bibitem{gu2019improving}
S.~Gu, Y.~Feng, and Q.~Liu.
\newblock Improving domain adaptation translation with domain invariant and specific information.
\newblock In {\em Proceedings of the 2019 Conference of the North American Chapter of the Association for Computational Linguistics: Human Language Technologies, Volume 1 (Long and Short Papers)}, pages 3081--3091, Minneapolis, Minnesota, 2019. NAACL.

\bibitem{hadsell2006dimensionality}
R.~Hadsell, S.~Chopra, and Y.~LeCun.
\newblock Dimensionality reduction by learning an invariant mapping.
\newblock In {\em 2006 IEEE computer society conference on computer vision and pattern recognition (CVPR'06)}, volume~2, pages 1735--1742, New York City, New York, 2006. IEEE.

\bibitem{han2022transpar}
Z.~Han, H.~Sun, and Y.~Yin.
\newblock Learning transferable parameters for unsupervised domain adaptation.
\newblock {\em IEEE Transactions on Image Processing}, 31:6424--6439, 2022.

\bibitem{he2016deep}
K.~He, X.~Zhang, S.~Ren, and J.~Sun.
\newblock Deep residual learning for image recognition.
\newblock In {\em Proceedings of the IEEE Conference on Computer Vision and Pattern Recognition}, pages 770--778, Las Vegas, Nevada, 2016. IEEE.

\bibitem{huang2017arbitrary}
X.~Huang and S.~Belongie.
\newblock Arbitrary style transfer in real-time with adaptive instance normalization.
\newblock In {\em Proceedings of the IEEE International Conference on Computer Vision}, pages 1501--1510, Venice, Italy, 2017. IEEE.

\bibitem{ionescu2013human3}
C.~Ionescu, D.~Papava, V.~Olaru, and C.~Sminchisescu.
\newblock Human3. 6m: Large scale datasets and predictive methods for 3d human sensing in natural environments.
\newblock {\em IEEE Transactions on Pattern Analysis and Machine Intelligence}, 36(7):1325--1339, 2013.

\bibitem{jiang2021regressive}
J.~Jiang, Y.~Ji, X.~Wang, Y.~Liu, J.~Wang, and M.~Long.
\newblock Regressive domain adaptation for unsupervised keypoint detection.
\newblock In {\em Proceedings of the IEEE/CVF Conference on Computer Vision and Pattern Recognition}, pages 6780--6789, Nashville, Tennessee, 2021. IEEE.

\bibitem{jin2022branch}
R.~Jin, J.~Zhang, J.~Yang, and D.~Tao.
\newblock Multibranch adversarial regression for domain adaptative hand pose estimation.
\newblock {\em IEEE Transactions on Circuits and Systems for Video Technology}, 32(9):6125--6136, 2022.

\bibitem{johnson2010clustered}
S.~Johnson and M.~Everingham.
\newblock Clustered pose and nonlinear appearance models for human pose estimation.
\newblock In {\em British Machine Vision Conference}, volume~2, page~5, Britain, 2010. Citeseer, BMVC.

\bibitem{kim2022unified}
D.~Kim, K.~Wang, K.~Saenko, M.~Betke, and S.~Sclaroff.
\newblock A unified framework for domain adaptive pose estimation.
\newblock In {\em European Conference on Computer Vision}, pages 603--620, Switzerland, 2022. Springer.

\bibitem{kingma2014adam}
D.~Kingma, L.~Ba, et~al.
\newblock Adam: A method for stochastic optimization.
\newblock 2015.

\bibitem{li2021synthetic}
C.~Li and G.~H. Lee.
\newblock From synthetic to real: Unsupervised domain adaptation for animal pose estimation.
\newblock In {\em Proceedings of the IEEE/CVF Conference on Computer Vision and Pattern Recognition}, pages 1482--1491, Nashville, Tennessee, 2021. IEEE.

\bibitem{li2021pose}
K.~Li, S.~Wang, X.~Zhang, Y.~Xu, W.~Xu, and Z.~Tu.
\newblock Pose recognition with cascade transformers.
\newblock In {\em Proceedings of the IEEE/CVF Conference on Computer Vision and Pattern Recognition}, pages 1944--1953, Nashville, Tennessee, 2021. IEEE.

\bibitem{lin2014microsoft}
T.-Y. Lin, M.~Maire, S.~Belongie, J.~Hays, P.~Perona, D.~Ramanan, P.~Doll{\'a}r, and C.~L. Zitnick.
\newblock Microsoft coco: Common objects in context.
\newblock In {\em European Conference on Computer Vision}, pages 740--755, Zurich, Switzerland, 2014. Springer.

\bibitem{long2015learning}
M.~Long, Y.~Cao, J.~Wang, and M.~Jordan.
\newblock Learning transferable features with deep adaptation networks.
\newblock In {\em International Conference on Machine Learning}, pages 97--105, Lille, France, 2015. PMLR.

\bibitem{luo2021rethinking}
Z.~Luo, Z.~Wang, Y.~Huang, L.~Wang, T.~Tan, and E.~Zhou.
\newblock Rethinking the heatmap regression for bottom-up human pose estimation.
\newblock In {\em Proceedings of the IEEE/CVF Conference on Computer Vision and Pattern Recognition}, pages 13264--13273, Nashville, Tennessee, 2021. IEEE.

\bibitem{mu2020learning}
J.~Mu, W.~Qiu, G.~D. Hager, and A.~L. Yuille.
\newblock Learning from synthetic animals.
\newblock In {\em Proceedings of the IEEE/CVF Conference on Computer Vision and Pattern Recognition}, pages 12386--12395, Seattle, Washington, 2020. IEEE.

\bibitem{peng2022multi}
Q.~Peng.
\newblock Multi-source and source-private cross-domain learning for visual recognition.
\newblock Master's thesis, Purdue University, 2022.

\bibitem{peng2023rain}
Q.~Peng, Z.~Ding, L.~Lyu, L.~Sun, and C.~Chen.
\newblock Rain: regularization on input and network for black-box domain adaptation.
\newblock In {\em Proceedings of the Thirty-Second International Joint Conference on Artificial Intelligence}, pages 4118--4126, Macau, China, 2023. IJCAI.

\bibitem{peng20243d}
Q.~Peng, B.~Planche, Z.~Gao, M.~Zheng, A.~Choudhuri, T.~Chen, C.~Chen, and Z.~Wu.
\newblock 3d vision-language gaussian splatting.
\newblock {\em arXiv preprint arXiv:2410.07577}, 2024.

\bibitem{peng2023source}
Q.~Peng, C.~Zheng, and C.~Chen.
\newblock Source-free domain adaptive human pose estimation.
\newblock In {\em Proceedings of the IEEE/CVF International Conference on Computer Vision}, pages 4826--4836, Paris, France, 2023. IEEE.

\bibitem{peng2024dual}
Q.~Peng, C.~Zheng, and C.~Chen.
\newblock A dual-augmentor framework for domain generalization in 3d human pose estimation.
\newblock In {\em Proceedings of the IEEE/CVF Conference on Computer Vision and Pattern Recognition}, pages 2240--2249, 2024.

\bibitem{pinyoanuntapong2023gaitsada}
E.~Pinyoanuntapong, A.~Ali, K.~Jakkala, P.~Wang, M.~Lee, Q.~Peng, C.~Chen, and Z.~Sun.
\newblock Gaitsada: Self-aligned domain adaptation for mmwave gait recognition.
\newblock In {\em 2023 IEEE 20th International Conference on Mobile Ad Hoc and Smart Systems (MASS)}, pages 218--226. IEEE, 2023.

\bibitem{raychaudhuri2023prior}
D.~S. Raychaudhuri, C.-K. Ta, A.~Dutta, R.~Lal, and A.~K. Roy-Chowdhury.
\newblock Prior-guided source-free domain adaptation for human pose estimation.
\newblock In {\em Proceedings of the IEEE/CVF International Conference on Computer Vision}, pages 14996--15006, Vancouver, Canada, 2023. IEEE.

\bibitem{saito2018maximum}
K.~Saito, K.~Watanabe, Y.~Ushiku, and T.~Harada.
\newblock Maximum classifier discrepancy for unsupervised domain adaptation.
\newblock In {\em Proceedings of the IEEE Conference on Computer Vision and Pattern Recognition}, pages 3723--3732, Salt Lake City, Utah, 2018. IEEE.

\bibitem{shi2022end}
D.~Shi, X.~Wei, L.~Li, Y.~Ren, and W.~Tan.
\newblock End-to-end multi-person pose estimation with transformers.
\newblock In {\em Proceedings of the IEEE/CVF Conference on Computer Vision and Pattern Recognition}, pages 11069--11078, New Orleans, Louisiana, 2022. IEEE.

\bibitem{stojanov2021domain}
P.~Stojanov, Z.~Li, M.~Gong, R.~Cai, J.~Carbonell, and K.~Zhang.
\newblock Domain adaptation with invariant representation learning: What transformations to learn?
\newblock {\em Advances in Neural Information Processing Systems}, 34:24791--24803, 2021.

\bibitem{sun2016deep}
B.~Sun and K.~Saenko.
\newblock Deep coral: Correlation alignment for deep domain adaptation.
\newblock In {\em European Conference on Computer Vision}, pages 443--450, Amsterdam, Netherlands, 2016. Springer.

\bibitem{sun2019deep}
K.~Sun, B.~Xiao, D.~Liu, and J.~Wang.
\newblock Deep high-resolution representation learning for human pose estimation.
\newblock In {\em Proceedings of the IEEE/CVF Conference on Computer Vision and Pattern Recognition}, pages 5693--5703, Long Beach, California, 2019. IEEE.

\bibitem{sun2017compositional}
X.~Sun, J.~Shang, S.~Liang, and Y.~Wei.
\newblock Compositional human pose regression.
\newblock In {\em Proceedings of the IEEE International Conference on Computer Vision}, pages 2602--2611, Venice, Italy, 2017. IEEE.

\bibitem{sun2018integral}
X.~Sun, B.~Xiao, F.~Wei, S.~Liang, and Y.~Wei.
\newblock Integral human pose regression.
\newblock In {\em Proceedings of the European Conference on Computer Vision (ECCV)}, pages 529--545, Munich, Germany, 2018. Springer.

\bibitem{tang2020unsupervised}
H.~Tang, K.~Chen, and K.~Jia.
\newblock Unsupervised domain adaptation via structurally regularized deep clustering.
\newblock In {\em Proceedings of the IEEE/CVF Conference on Computer Vision and Pattern Recognition}, pages 8725--8735, Seattle, Washington, 2020. IEEE.

\bibitem{tarvainen2017mean}
A.~Tarvainen and H.~Valpola.
\newblock Mean teachers are better role models: Weight-averaged consistency targets improve semi-supervised deep learning results.
\newblock {\em Advances in Neural Information Processing Systems}, 30:1195--1204, 2017.

\bibitem{toshev2014deeppose}
A.~Toshev and C.~Szegedy.
\newblock Deeppose: Human pose estimation via deep neural networks.
\newblock In {\em Proceedings of the IEEE Conference on Computer Vision and Pattern Recognition}, pages 1653--1660, Columbus, Ohio, 2014. IEEE.

\bibitem{tzeng2014deep}
E.~Tzeng, J.~Hoffman, N.~Zhang, K.~Saenko, and T.~Darrell.
\newblock Deep domain confusion: Maximizing for domain invariance.
\newblock {\em arXiv preprint arXiv:1412.3474}, 2014.

\bibitem{varol2017learning}
G.~Varol, J.~Romero, X.~Martin, N.~Mahmood, M.~J. Black, I.~Laptev, and C.~Schmid.
\newblock Learning from synthetic humans.
\newblock In {\em Proceedings of the IEEE Conference on Computer Vision and Pattern Recognition}, pages 109--117, Honolulu, Hawaii, 2017. CVPR.

\bibitem{wu2021heter}
H.~Wu, H.~Zhu, Y.~Yan, J.~Wu, Y.~Zhang, and M.~K. Ng.
\newblock Heterogeneous domain adaptation by information capturing and distribution matching.
\newblock {\em IEEE Transactions on Image Processing}, 30:6364--6376, 2021.

\bibitem{xiao2018simple}
B.~Xiao, H.~Wu, and Y.~Wei.
\newblock Simple baselines for human pose estimation and tracking.
\newblock In {\em Proceedings of the European Conference on Computer Vision (ECCV)}, pages 466--481, Munich, Germany, 2018. Springer.

\bibitem{xin2024vmt}
Y.~Xin, J.~Du, Q.~Wang, Z.~Lin, and K.~Yan.
\newblock Vmt-adapter: Parameter-efficient transfer learning for multi-task dense scene understanding.
\newblock In {\em Proceedings of the AAAI Conference on Artificial Intelligence}, volume~38, pages 16085--16093, 2024.

\bibitem{xin2024mmap}
Y.~Xin, J.~Du, Q.~Wang, K.~Yan, and S.~Ding.
\newblock Mmap: Multi-modal alignment prompt for cross-domain multi-task learning.
\newblock In {\em Proceedings of the AAAI Conference on Artificial Intelligence}, volume~38, pages 16076--16084, 2024.

\bibitem{xin2024vpetl}
Y.~Xin, S.~Luo, X.~Liu, Y.~Du, H.~Zhou, X.~Cheng, C.~L. Lee, J.~Du, H.~Wang, M.~Chen, et~al.
\newblock V-petl bench: A unified visual parameter-efficient transfer learning benchmark.
\newblock In {\em The Thirty-eight Conference on Neural Information Processing Systems Datasets and Benchmarks Track}.

\bibitem{xu2019larger}
R.~Xu, G.~Li, J.~Yang, and L.~Lin.
\newblock Larger norm more transferable: An adaptive feature norm approach for unsupervised domain adaptation.
\newblock In {\em Proceedings of the IEEE/CVF International Conference on Computer Vision}, pages 1426--1435, Seoul, South Korea, 2019. IEEE.

\bibitem{zhang2020distribution}
F.~Zhang, X.~Zhu, H.~Dai, M.~Ye, and C.~Zhu.
\newblock Distribution-aware coordinate representation for human pose estimation.
\newblock In {\em Proceedings of the IEEE/CVF Conference on Computer Vision and Pattern Recognition}, pages 7093--7102, Seattle, Washington, 2020. IEEE.

\bibitem{zhao2020h3d}
Z.~Zhao, T.~Wang, S.~Xia, and Y.~Wang.
\newblock Hand-3d-studio: A new multi-view system for 3d hand reconstruction.
\newblock In {\em ICASSP 2020-2020 IEEE International Conference on Acoustics, Speech and Signal Processing (ICASSP)}, pages 2478--2482, Barcelona, Spain, 2020. IEEE.

\bibitem{zheng2023feater}
C.~Zheng, M.~Mendieta, T.~Yang, G.-J. Qi, and C.~Chen.
\newblock Feater: An efficient network for human reconstruction via feature map-based transformer.
\newblock In {\em Proceedings of the IEEE/CVF Conference on Computer Vision and Pattern Recognition}, 2023.

\bibitem{zheng20213d}
C.~Zheng, S.~Zhu, M.~Mendieta, T.~Yang, C.~Chen, and Z.~Ding.
\newblock 3d human pose estimation with spatial and temporal transformers.
\newblock In {\em Proceedings of the IEEE/CVF international conference on computer vision}, pages 11656--11665, 2021.

\bibitem{zhu2019aligning}
Y.~Zhu, F.~Zhuang, and D.~Wang.
\newblock Aligning domain-specific distribution and classifier for cross-domain classification from multiple sources.
\newblock In {\em Proceedings of the AAAI Conference on Artificial Intelligence}, volume~33, pages 5989--5996, 2019.

\bibitem{zhuang2020comprehensive}
F.~Zhuang, Z.~Qi, K.~Duan, D.~Xi, Y.~Zhu, H.~Zhu, H.~Xiong, and Q.~He.
\newblock A comprehensive survey on transfer learning.
\newblock {\em Proceedings of the IEEE}, 109(1):43--76, 2020.

\bibitem{zimmermann2017learning}
C.~Zimmermann and T.~Brox.
\newblock Learning to estimate 3d hand pose from single rgb images.
\newblock In {\em Proceedings of the IEEE International Conference on Computer Vision}, pages 4903--4911, Honolulu, Hawaii, 2017. IEEE.

\bibitem{zimmermann2019freihand}
C.~Zimmermann, D.~Ceylan, J.~Yang, B.~Russell, M.~Argus, and T.~Brox.
\newblock Freihand: A dataset for markerless capture of hand pose and shape from single rgb images.
\newblock In {\em Proceedings of the IEEE/CVF International Conference on Computer Vision}, pages 813--822, Long Beach, California, 2019. IEEE.

\end{thebibliography}
}

\newpage
\section{Supplementary Material Overview}

The supplementary material is organized into the following sections:

\begin{itemize}
    \item Section \ref{sec:param}: Parameter sensitivity analysis. 
    \item Section \ref{sec:ab}: Additional ablation study on FreiHand and LSP datasets.
    \item Section \ref{sec:ab-dl}: Additional ablation of discrepancy loss terms on FreiHand and LSP datasets.
    \item Section \ref{sec:ab-func}: Ablation of varied loss functions to measure discrepancy.
    \item Section \ref{sec:gen}: Domain generalization to unseen domains based on models trained on domain adaptation tasks.
    \item Section \ref{sec:coco}:
    Qualitative results on the unseen dataset COCO based on models trained on domain adaptation tasks.
\end{itemize}

\section{Parameter Sensitivity Analysis}
\label{sec:param}

\begin{figure}[!ht]
  \centering
  \includegraphics[width=1\linewidth]{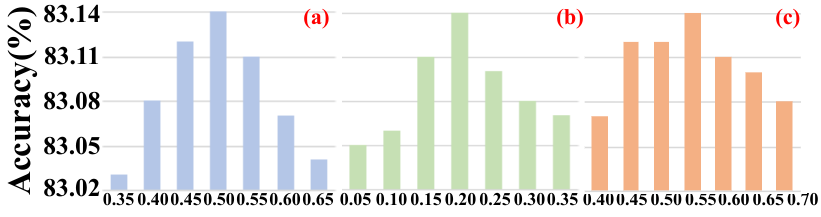}
  \caption{Parameter Analysis on RHD$\rightarrow$H3D (best viewed in color). \textbf{a}: Analysis on $\alpha$. \textbf{b}: Analysis on $\beta$. \textbf{c}: Analysis on $\gamma$. 
   }
   \label{fig:param_hand}
\end{figure}

\begin{figure}[!ht]
  \centering
  \includegraphics[width=1\linewidth]{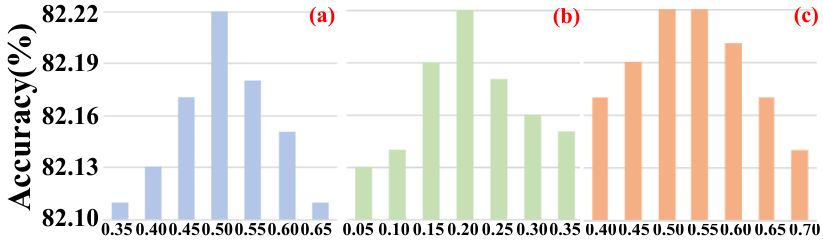}
  \caption{Analysis on SURREAL$\rightarrow$Human3.6M (best viewed in color). \textbf{a}: Analysis on $\alpha$. \textbf{b}: Analysis on $\beta$. \textbf{c}: Analysis on $\gamma$. 
   }\vspace{-10pt}
   \label{fig:param_human}
\end{figure}

We use RHD$\rightarrow$H3D and SURREAL$\rightarrow$ Human3.6M to illustrate the sensitivity of $\alpha_{1}$ and $\alpha_{2}$, $\beta$, and $\gamma$ in the main paper. Since the terms associated with $\alpha_{1}$ and $\alpha_{2}$ are very similar (in order to warm up two heads), we integrate them into $\alpha$. Fig. \ref{fig:param_hand} and Fig. \ref{fig:param_human} exhibit the results under different parameter settings. From it, we observe that the decreases from different parameters are limited to $0.12\%$ for $\alpha$, $\beta$, and $\gamma$. Therefore, our method shows stable performance over hyperparameters.

\section{Additional Ablation Study}
\label{sec:ab}

In this section, we provide additional ablation study on the two tasks RHD$\rightarrow$FreiHand and the SURREAL$\rightarrow$LSP, and the results are shown in Table \ref{tab:rf-ab} and Table \ref{tab:sl-ab}. There are three components \textbf{Intermediate Domain Framework} (\textbf{IDF}), \textbf{Discrepancy Loss} (\textbf{DL}), and \textbf{Reweighting} (\textbf{RW}). On the RHD $\rightarrow$ FreiHand task, applying only `Inter' to `IDF' leads to an increase of $4.0\%$, and using only `IDF w/ Spec' results in an improvement of $3.2\%$. In contrast, when `DL' is formed, the performance boost becomes substantial at $9.6\%$. The results show that the process of learning invariant representations relies on both aggregating domain-invariant features and segregating domain-specific features, and these two processes can help each other get better. 

\begin{table}[!ht]
    \scriptsize
    \centering
    \caption{Ablation Study on RHD $\rightarrow$ FreiHand Task}
    \resizebox{0.9\linewidth}{!}{%
    \begin{tabular}{ccccccccc}
         \toprule
         Method &  MCP &  PIP  &  DIP & Fin & All \\
         \hline
         {Baseline} & 40.3 & 55.8 & 57.9 & 52.9 & 50.6\\
         {Baseline w/ DL} & 41.1 & 57.6 & 58.5 & 54.0 & 53.4\\
         \hline

         {AIDF} & 40.5 & 55.6 & 57.8 & 53.0 & 50.6\\
         {AIDF w/ Inter} & 41.7 & 56.8 & 59.9 & 54.4 & 52.1\\
         {AIDF w/ Spec} & 41.1 & 55.9 & 58.6 & 54.0 & 51.7\\
         {AIDF w/ DL} & 43.6 & 58.1 & 60.9 & 58.2 & 55.4\\

         \hline
         {IDF} & 40.4 & 55.5 & 57.9 & 53.2 & 50.7\\
         {IDF w/ Inter} & 42.8 & 60.4 & 62.5 & 56.9 & 54.7\\
         {IDF w/ Spec} & 42.4 & 58.5 & 59.7 & 55.1 & 53.9\\
         {IDF w/ DL (Ours)} & \textbf{46.0} & \textbf{65.2} & \textbf{69.5} & \textbf{63.7} & \textbf{61.3}\\  
         \bottomrule
    \end{tabular}%
    }
    \vspace{-2pt}
\label{tab:rf-ab}
\end{table}

\begin{table}[!ht]
    \scriptsize
    \centering
    \caption{Ablation Study on SURREAL $\rightarrow$ LSP Task}
    \resizebox{0.97\linewidth}{!}{%
    \begin{tabular}{ccccccccc}
          \toprule
          Method &  Sld & Elb & Wrist & Hip & Knee & Ankle & All \\
          \hline
          {Baseline} & 62.8 & 77.0 & 71.3 & 81.4 & 80.5 & 75.6 & 74.9\\
          {Baseline w/ DL} & 64.0 & 77.7 & 72.5 & 82.0 & 81.6 & 77.1 & 76.5\\
          \hline

          {AIDF} & 63.2 & 77.1 & 71.5 & 81.0 & 80.6 & 75.4 & 74.8\\
          {AIDF w/ Inter} &  64.1 & 78.2 & 73.3 & 82.4 & 82.2 & 77.6 & 76.9\\
          {AIDF w/ Spec} & 63.4 & 77.5 & 72.0 & 81.7 & 81.4 & 76.0 & 75.5\\
          {AIDF w/ DL}  & 66.6 & 80.8 & 75.4 & 84.2 & 83.0 & 80.9 & 81.1\\

          \hline
          {IDF} & 62.9 & 77.2 & 71.7 & 81.4 & 80.5 & 75.9 & 74.9\\
          {IDF w/ Inter} & 65.9 & 80.8 & 74.0 & 84.1 & 83.3 & 81.6 & 80.6\\
          {IDF w/ Spec} & 64.4 & 79.3 & 73.5 & 82.8 & 82.0 & 79.7 & 78.3\\
          {IDF w/ DL (Ours)} &  \textbf{72.1} & \textbf{86.4} & \textbf{85.2} & \textbf{87.7} & \textbf{87.0} & \textbf{86.2} & \textbf{84.8}\\
         \bottomrule
    \end{tabular}%
    }
    \vspace{-2pt}
\label{tab:sl-ab}
\end{table}

\section{Additional Ablation of Discrepancy Loss Terms}
\label{sec:ab-dl}

In this section, we present additional ablation of discrepancy loss terms on the RHD$\rightarrow$FreiHand task and the SURREAL$\rightarrow$LSP task, and the results are listed in Table \ref{tab:rf-ab-dl} and Table \ref{tab:sl-ab-dl}. As described in the main manuscript, three terms $r_1$, $r_2$, and $r_3$ are considered in the proposed method. It is noticeable that $r_2$ plays a more important role than $r_3$, since the removal of $r_2$ leads to a decrease of $3.6\%$ in RHD$\rightarrow$FreiHand and $2.7\%$ in SURREAL$\rightarrow$LSP, while that of $r_3$ results in a reduction of $1.7\%$ in RHD$\rightarrow$FreiHand and $1.5\%$ in SURREAL$\rightarrow$LSP.

\begin{table}[!ht]
    \scriptsize
    \centering
    \caption{Ablation of discrepancy loss terms on RHD $\rightarrow$ FreiHand Task}
    \resizebox{0.9\linewidth}{!}{%
    \begin{tabular}{ccccccccc}
          \toprule
          Method &  MCP &  PIP  &  DIP & Fin & All \\
         \hline
         {$r_1$} & 41.6 & 62.7 & 65.8 & 60.2 & 56.9\\
         {$r_2$} & 40.9 & 61.1 & 64.0 & 59.7 & 55.2\\
         {$r_3$} & 40.4 & 61.3 & 64.4 & 60.9 & 57.8\\
         {$r_1$ \& $r_2$} & 44.1 & 65.3 & 68.7 & 62.1 & 59.0\\
         {$r_1 $ \& $r_3$} & 42.6 & 64.6 & 68.3 & 61.5 & 57.7\\
         {$r_2 $ \& $r_3$} & 43.5 & 64.9 & 68.9 & 61.8 & 58.6\\
         {$r_1$ \& $r_2$ \& $r_3$} & 46.0 & 65.2 & 69.5 & 63.7 & 61.3 \\    
         \bottomrule
    \end{tabular}%
    }
    \vspace{-2pt}
\label{tab:rf-ab-dl}
\end{table}

\begin{table}[!ht]
    \scriptsize
    \centering
    \caption{Ablation of discrepancy loss terms on SURREAL $\rightarrow$ LSP Task}
    \resizebox{0.97\linewidth}{!}{%
    \begin{tabular}{ccccccccc}
          \toprule
          Method &  Sld & Elb & Wrist & Hip & Knee & Ankle & All \\
         \hline
         {$r_1$} & 67.9 & 82.2 & 81.8 & 84.7 & 84.3 & 80.1 & 80.6 \\
         {$r_2$} & 65.3 & 80.7 & 80.0 & 82.5 & 81.6 & 78.8 & 78.2\\
         {$r_3$} & 66.4 & 81.3 & 80.9 & 83.7 & 82.2 & 79.0 & 78.8\\
         {$r_1$ \& $r_2$} & 71.0 & 85.0 & 83.3 & 84.1 & 85.5 & 84.7 & 83.2\\
         {$r_1$ \& $r_3$} & 70.6 & 83.8 & 81.3 & 82.6 & 84.8 & 83.7 & 81.9\\
         {$r_2$ \& $r_3$} & 70.7 & 84.3 & 81.9 & 83.5 & 85.4 & 84.2 & 82.5\\
         {$r_1$ \& $r_2$ \& $r_3$}  & 72.1 & 86.4 & 85.2 & 87.7 & 87.0 & 86.2 & 84.8\\    
         \bottomrule
    \end{tabular}%
    }
    \vspace{-2pt}
\label{tab:sl-ab-dl}
\end{table}

\section{Ablation of varied loss functions}
\label{sec:ab-func}

In our proposed method, MMD loss \cite{zhu2019aligning} is applied to measure discrepancy. Unlike KL divergence, which measures the distance between two distributions, and MSE loss, which gauges the spatial difference between two heatmaps, the MMD loss is a specific metric crafted for assessing the distance between two hypotheses \cite{zhu2019aligning,gretton2012kernel}. \textbf{This is particularly relevant to our proposed multi-branch (multi-hypotheses) framework.} Furthermore, we substitute the MMD loss in $\mathcal{L}_{dl}$ with KL divergence and MSE loss for empirical comparisons. The results are presented in Tab. \ref{tab:ab-loss}, showing the superiority of MMD over MSE and KL Divergence.

\begin{table}[!ht]
    \scriptsize
    \centering
    \caption{Ablation of loss functions on RHD $\rightarrow$ H3D task}
    \resizebox{1.0\linewidth}{!}{%
    \begin{tabular}{ccccccccc}
          \toprule
          Method &  MCP &  PIP  &  DIP & Fin & All \\
         \hline
         {MSE} & 82.7 & 75.9 & 71.0 & 62.2 & 73.5\\
         {KL Divergence} & 83.3 & 80.4 & 75.7 & 63.9 & 77.8 \\
         {MMD (Ours)} & \textbf{89.6} & \textbf{88.5} & \textbf{82.4} & \textbf{73.0} & \textbf{83.5}\\    
         \bottomrule
    \end{tabular}%
    }
\label{tab:ab-loss}
\end{table}

\section{Generalization to Unseen Domains}
\label{sec:gen}

Following prior works \cite{li2021synthetic,kim2022unified}, we also conduct experiments on the generalization to unseen domains. For hand pose estimation, we use models adapted in the RHD$\rightarrow$H3D task and evaluate their performances on the validation set of FreiHand, as shown in Table \ref{tab:frei_gene}. For human pose estimation, we use models adapted in the SURREAL$\rightarrow$LSP task and evaluate their performance on Human3.6M, as shown in Table \ref{tab:h36m_gene}.

\begin{table}[!ht]
    \scriptsize
    \centering
    \caption{Domain Generalization on FreiHand}
    \resizebox{0.9\linewidth}{!}{%
    \begin{tabular}{ccccccccc}
          \toprule
          Method &  MCP &  PIP  &  DIP & Fin & All \\
         \hline
         {Source-only} & 34.9 & 48.7 & 52.4 & 48.5 & 45.8\\
         {Oracle} & 92.8 & 90.3 & 87.7 & 78.5 & 87.2\\
         \hline
         CC-SSL \cite{mu2020learning}  & 34.3 & 46.3 & 48.4 & 44.4 & 42.6 \\
         MDAM \cite{li2021synthetic}  & 29.6 & 46.6 & 50.0 & 45.3 & 42.2\\
         RegDA \cite{jiang2021regressive}  & 37.8 & 51.8 & 53.2 & 47.5 & 46.9\\
         UniFrame \cite{kim2022unified}  & 35.6 & 52.3 & \textbf{55.4} & \textbf{50.6} & 47.1\\
         \hline
         {Ours} & \textbf{38.0} & \textbf{52.5} & 54.8 & \textbf{50.6} & \textbf{47.4} \\    
         \bottomrule
    \end{tabular}%
    }
    \vspace{-2pt}
\label{tab:frei_gene}
\end{table}

\begin{table}[!ht]
    \scriptsize
    \centering
    \caption{Domain Generalization on Human3.6M}
    \resizebox{0.97\linewidth}{!}{%
    \begin{tabular}{ccccccccc}
          \toprule
          Method &  Sld & Elb & Wrist & Hip & Knee & Ankle & All \\
         \hline
         {Source-only} & 51.5 & 65.0 & 62.9 & 68.0 & 68.7 & 67.4 & 63.9\\
         {Oracle} & 95.3 & 91.8 & 86.9 & 95.6 & 94.1 & 93.6 & 92.9\\
         \hline
         CC-SSL \cite{mu2020learning} & 52.7 & 76.9 & 63.1 & 31.6 & 75.7 & 72.9 & 62.2 \\
         MDAM \cite{li2021synthetic}  & 54.4 & 75.3 & 62.1 & 21.6 & 70.4 & 69.2 & 58.8\\
         RegDA \cite{jiang2021regressive}  & 76.9 & 80.2 & 69.7 & 52.0 & 80.3 & 80.0 & 73.2\\
         UniFrame \cite{kim2022unified} & 77.0 & \textbf{85.9} & 73.8 & 47.6 & \textbf{80.7} & 80.6 & 74.3\\
         \hline
         {Ours} & \textbf{77.5} & 82.9 & \textbf{74.1} & \textbf{53.3} & 80.2 & \textbf{80.8} & \textbf{74.7}  \\    
         \bottomrule
    \end{tabular}%
    }
    \vspace{-2pt}
\label{tab:h36m_gene}
\end{table}

From these two tables, we can see that our model still outperforms other approaches for a lead of $0.3\%$ on FreiHand and $0.4\%$ on Human3.6M, which is not as large as that of the adaptation tasks presented in the main paper. This is partly because our model is designed for adaptation problems, not for generalization settings specifically. Nevertheless, our method still achieves the best results in these generalization experiments as compared to the state-of-the-art, demonstrating its superiority. 

\section{Qualitative Results on the Unseen Dataset COCO}
\label{sec:coco}

\begin{figure*}[!ht]
  \centering
   \includegraphics[width=0.99\linewidth]{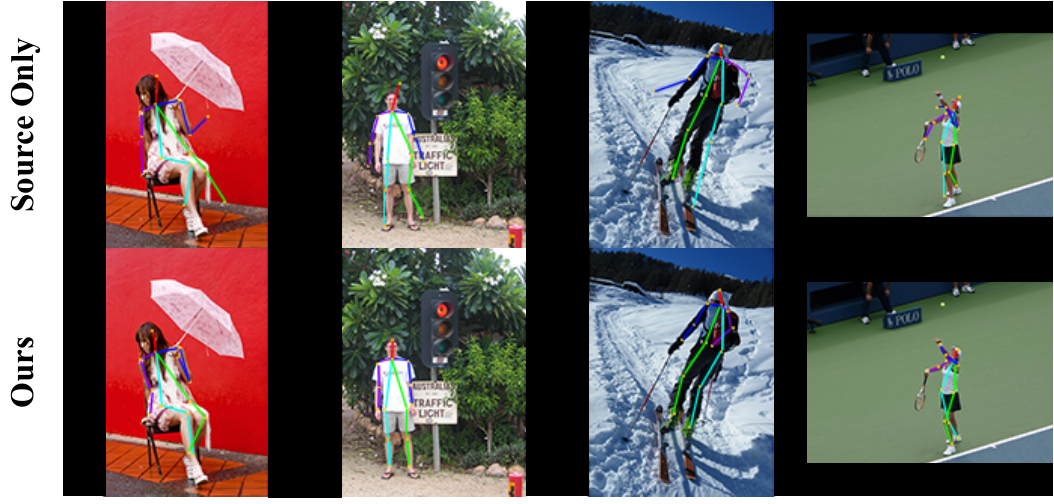}
   \caption{ Qualitative results on the COCO dataset. Here we compare our method trained from SURREAL $\rightarrow$ LSP with the source-only model.
   }
   \label{fig:coco}
\end{figure*}

In this section, we present qualitative visualizations on the COCO dataset \cite{lin2014microsoft}, which was not included in our previous human dataset studies due to differences in annotation methods. While we cannot provide quantitative generalization results due to these discrepancies, we believe that our qualitative results are still meaningful. To conduct our analysis, we utilized the model trained on the SURREAL $\rightarrow$ LSP task in combination with the pre-trained SURREAL model, and the resulting visualizations are shown in Fig. \ref{fig:coco}. These results demonstrate that our model is capable of generalizing to an unseen dataset, suggesting that the adaptation process has a positive impact on the model's generalization ability.

\end{document}